%% file: paper_template.tex
\documentclass[conference]{IEEEtran}
\usepackage{times}

\usepackage[numbers]{natbib}
\usepackage{multicol}
 \usepackage[hidelinks]{hyperref} 
 
\usepackage{amsmath} 
\usepackage{amssymb}
\usepackage{bm}
\usepackage{url}
\usepackage{graphicx,color,psfrag}
\usepackage{caption}
\usepackage{subcaption}
\usepackage{framed}
\usepackage{url}

\begin{document}

\title{Incremental Nonlinear Dynamic Inversion based Optical Flow Control for Flying Robots: \\An Efficient Data-driven Approach}

\author{\authorblockN{Hann Woei Ho and Ye Zhou*}
\authorblockA{School of Aerospace Engineering, Engineering Campus, Universiti Sains Malaysia, Malaysia\\
Faculty of Aerospace Engineering, Delft University of Technology, The Netherlands\\
The authors contributed equally to this work.\\
Email: aehannwoei@usm.my, zhouye@usm.my  *Corresponding author}
}


%
\onecolumn This paper has been accepted for publication at the Robotics: Science and Systems conference (RSS) 2023.
\twocolumn
\maketitle

\begin{abstract}
This paper presents a novel approach for optical flow control of Micro Air Vehicles (MAVs). The task is challenging due to the nonlinearity of optical flow observables. Our proposed Incremental Nonlinear Dynamic Inversion (INDI) control scheme incorporates an efficient data-driven method to address the nonlinearity. It directly estimates the inverse of the time-varying control effectiveness in real-time, eliminating the need for the constant assumption and avoiding high computation in traditional INDI. This approach effectively handles fast-changing system dynamics commonly encountered in optical flow control, particularly height-dependent changes. We demonstrate the robustness and efficiency of the proposed control scheme in numerical simulations and also real-world flight tests: multiple landings of an MAV on a static and flat surface with various tracking setpoints, hovering and landings on moving and undulating surfaces. Despite being challenged with the presence of noisy optical flow estimates and the lateral and vertical movement of the landing surfaces, the MAV is able to successfully track or land on the surface with an exponential decay of both height and vertical velocity at almost the same time, as desired. Video demos can be found at \url{https://youtu.be/i2xFJNGhfxs}.
\end{abstract}

\IEEEpeerreviewmaketitle

\section{Introduction}
Recent advancements in the development of small flying robots have opened up their potential for a wide range of indoor applications due to their ability to navigate in confined spaces and lower risk of harm when operating near users \cite{soria2022swarms,fridovich2020confidence}. Despite this, achieving full autonomy in these robots remains a challenge due to their limited payload capacities and computational resources.

Extensive research has been conducted to design solutions inspired by biological systems, such as tiny flying insects, which face similar limitations but are still capable of performing complex tasks. These insects heavily rely on visual information for navigation and control, and it is believed by biologists that their movements may be guided by optical flow cues \cite{wang2022recovery, collett2002insect}.

Therefore, the use of optical flow in Micro Air Vehicles (MAVs) has been a subject of growing interest in recent years due to its computational efficiency. The optical flow, which refers the apparent motion of features in a visual scene, can be used to provide rich information of the surroundings for tasks, such as hovering, landing, and obstacle avoidance \cite{yu2022visual, mahlknecht2022exploring, falanga2020dynamic}. The low computational demands and ability to function in GPS-denied environments make optical flow methods particularly well-suited for use in MAVs.

\begin{figure}[!t]
	\centering
	\input{intro.tex}
	\caption{Optical flow landing of an MAV on an undulating surface while moving laterally using our proposed INDI control scheme. The lateral movement of the MAV was deliberately generated by tracking a constant lateral flow $\vartheta_y^*$ to excite the challenging vertical movement of the landing surface as perceived by its bottom camera. The controller also tracked a constant flow divergence $\vartheta_z^*$, resulting both relative height $\tilde{d}_z$ and velocity $\tilde{v}_z$ decay exponentially to zero simultaneously. All vision and control algorithms were run onboard the MAV.}
	\label{fig:intro}
\end{figure}
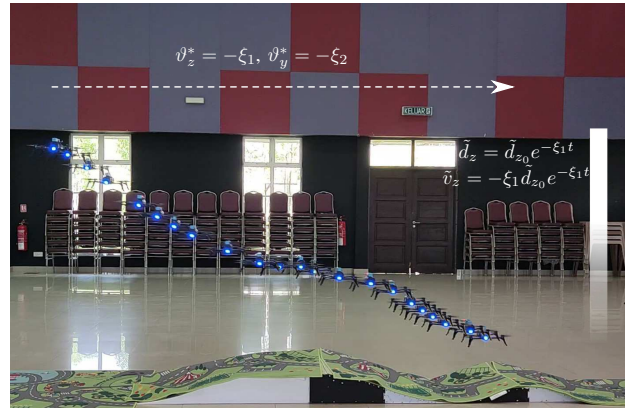

However, the use of optical flow methods poses significant control problems. The optical flow observables are highly nonlinear, being represented as the ratios of the velocities to the height of the MAV, which makes the controller design more challenging. Additionally, the estimation of optical flow observables can be affected by various sources of noise and error, such as sensor noise, motion blur, and lighting changes, further exacerbating the control challenges. 

Traditional control approaches, such as PID control and model-based Lyapunov methods \cite{ruffier2015optic, herisse2011landing}, which are commonly used in optical flow control for MAVs. These methods are typically designed for systems with linear dynamics and known models, making them sensitive to parameter uncertainties and model errors. This can lead to poor performance and instability of the control system when using optical flow observables as its control input \cite{ho2018adaptive}. 

As an alternative, adaptive approaches, such as bio-inspired strategies and machine learning \cite{de2022accommodating, o2022neural}, can be used to improve the performance of the control system in the presence of nonlinearities. These approaches are based on the idea of using feedback from the system to adapt the control strategy in real-time, making them more robust to unknown dynamics and nonlinearities. However, these methods can be complex to implement and require significant computational resources. 

Nonlinear control approaches, such as PID scheduling \cite{kendoul2014four, cesetti2010vision}, which involves the tuning of multiple parameters, can also be employed to deal with nonlinear optical flow control. However, the implementation of these methods can be sophisticated, as the parameters need to be carefully tuned to achieve good performance. Furthermore, the use of PID scheduling in optical flow control for MAVs can be problematic as it relies heavily on the accurate knowledge of the system dynamics and parameters, which are often unknown or uncertain in real-world applications. Additionally, the performance of PID scheduling can be sensitive to parameter variations and model errors, which can lead to poor performance and instability of the control system in the presence of nonlinearities. 

In contrast, the Incremental Nonlinear Dynamic Inversion (INDI) method has been shown to be a suitable approach for nonlinear systems with unknown models and nonlinear output \cite{zhou2021extended}. INDI is a nonlinear control method that utilizes the inverse of the system dynamics to cancel out the nonlinearities present in the system. By using this inverse, INDI can provide a linear control input that can be used to stabilize the system. It is also robust to parameter uncertainties and model errors, making it an effective method for handling nonlinear optical flow control in MAVs. Furthermore, INDI is a computationally efficient method that can be implemented in real-time, making it a viable option for practical applications.

The current implementation of INDI utilizes a constant control effectiveness matrix $G$, which is determined by the system dynamics \cite{steffensen2022nonlinear, smeur2018cascaded}. However, when it comes to optical flow stabilization, landing, or tracking in MAVs, the system dynamics are dependent on the height of the vehicle above the ground. As the height of the vehicle changes, the dynamics of the system also change. Therefore, a constant $G$ matrix is inappropriate for optical flow control in MAVs. To address this issue, it would be beneficial to develop an INDI method that uses a $G$ matrix that changes over time. This would enable the control system to adapt to the changing dynamics of the system, resulting in improved performance and stability.

Therefore, the identification of the time-varying $G$ matrix is essential for INDI-based optical flow control. However, the calculation of the $G$ matrix is a computationally intensive task and can be challenging in real-time applications. To address this issue, we propose a novel INDI control scheme that directly identifies its inverse, i.e., $G^\dagger=G^{-1}$. In other words, it could reduce the computation required for the control increment, resulting in a significant reduction in computation time. Furthermore, this approach eliminates the error introduced by the inversion calculation and thus improves the accuracy of the control system. 

Figure \ref{fig:intro} illustrates one of the flight tests conducted in this paper using our proposed INDI control scheme, i.e., the optical flow landing of an MAV on an undulating surface while translating. We demonstrate the success of the tracking of the desired constant optical flow setpoints, i.e., $\vartheta_z^*$ and $\vartheta_y^*$, with the INDI control, allowing the desired exponential decay of both relative height $\tilde{d}_z$ and velocity $\tilde{v}_z$ to zero in a nearly simultaneous manner while moving laterally. The lateral movement of the MAV was intentionally generated to stimulate the complex vertical movement of the landing surface observed by the onboard downward-facing camera to validate the robustness of the control. 

\section{Problem Formulation}
This section describes the dynamics of an MAV and its observed optical flow signal during the landing process onto a moving platform. 
The attitude of the MAV is stabilized through an inner control loop, which is designed to be accurate and responsive in its operation. 

\subsection{The position dynamics of an MAV}

\begin{figure}[!t]
	\centering
	\input{uav_body_axis.tex}
	\caption{MAV body ($O^bX^bY^bZ^b$) and inertial ($O^IX^IY^IZ^I$) reference frames. $O^b$ is fixed at the center of gravity of the MAV, $X^b$ points forward, $Y^b$ is port, and $Z^b$ points upward. The inertial reference frame is located on the ground and follows the North-East-Up ($X^I-Y^I-Z^I$) system.}
	\label{fig:uav_body_axis}
\end{figure}
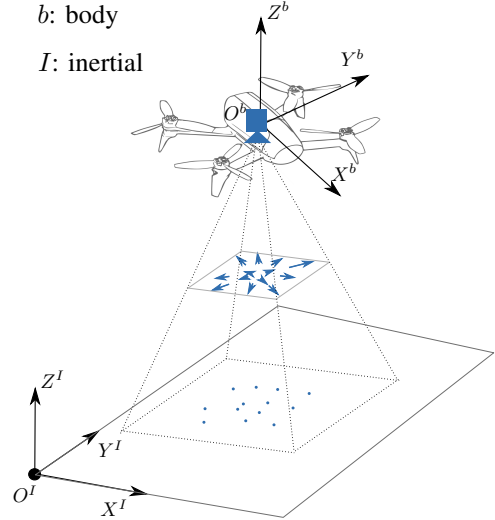

Figure \ref{fig:uav_body_axis} defines the body reference frame of an MAV ($O^bX^bY^bZ^b$) and the inertial reference frame ($O^IX^IY^IZ^I$). A camera is attached to the body of the MAV and looking downward. Thus, the camera reference frame ($O^cX^cY^cZ^c$) is assumed to align with the body reference frame. In the rest of the paper, the variables are in the inertial reference frame unless indicated with the superscripts $b$ and $c$ only to avoid confusion. 

The position $\bm{d}$ and velocity $\bm{v}$ of the MAV have the dynamics given by Newton's second law:
\begin{align}
\label{eq:positionDynamics1}
\dot{\bm{d}} &= \bm{v},\\
\label{eq:positionDynamics2}
\dot{\bm{v}} &= \bm{g} + \frac{1}{M} \bm{T} (T, \bm{\omega}) 
+ \frac{1}{M} \bm{\delta}(\bm{v}, \bm{\epsilon}),
\end{align}
where $\bm{g}$ is gravity, $M$ is the mass, $\bm{T}$ is the thrust generated by the rotors, and it is a function of the total thrust in $Z_b$ direction $T$ and the attitude in Euler angles $\bm{\omega}=[\theta, \phi, \psi]$. 
The term $\bm{\delta}$ represents the aerodynamic forces due to the velocity $\bm{v}$ and other factors $\bm{\epsilon}$, such as wind, acting on the MAV. 

In the tracking moving platform and landing tasks, we can assume that the changes in the yaw angle $\psi$ is small, and we can align the initial MAV position so that the yaw angle $\psi$ can be neglected. The position dynamics in Eq. (\ref{eq:positionDynamics2}) becomes
\begin{equation}
\label{eq:positionDynamicsExp}
\dot{\bm{v}} = \begin{bmatrix}
0 \\
0 \\
-g
\end{bmatrix} + \frac{1}{M} \begin{bmatrix}
\cos \phi \sin \theta \\
-\sin \phi \\
\cos \phi \cos \theta
\end{bmatrix} T + \frac{1}{M} \bm{\delta}(\bm{v}, \bm{\epsilon}).
\end{equation}
The attitude of the MAV is controlled by an inner control loop, thus, the input of the position control is $\bm{u}=[\theta, \phi, T]^T$.

\subsection{The observed optical flow signal}

The camera attached to the MAV faces downward and perceives optical flow used for the landing on a moving platform. 
To simplify the system so that the underlying control principles can be explained more clearly, this paper focuses on the problem of tracking and landing on a flat platform with slow, translational degrees of freedom. Its position and velocity vectors are $\bm{d}^r$ and $\bm{v}^r$, where the superscript $r$ stands for the reference platform. 
During movements, the camera will capture the optical flow, which is the velocity of the features on the platform beneath the camera. It is proportional to the ratio of the relative velocity to the height above the platform $d_z-d_z^r$. 
In this configuration, the camera is able to capture the three-dimensional flow field of the features beneath it, constituting the output vector of this problem: $\bm{y} = [\vartheta_x, \vartheta_y, \vartheta_z]^T$.

The optical flow captured in the $X$- and $Y$-axes represents the lateral flow of features, which describes the velocity of the features in the image plane. These measurements can be formulated as the ratio of the relative lateral velocity to the height above the platform:
\begin{equation}
\label{eq:opticalFlowX}
\vartheta_x = c_x \frac{v_x-v_x^r}{{d_z-d_z^r}},
\end{equation}
\begin{equation}
\label{eq:opticalFlowY}
\vartheta_y = c_y \frac{v_y-v_y^r}{{d_z-d_z^r}},
\end{equation}
where $c_x$ and $c_y$ are unknown proportionality constants. These constants are intrinsic properties of the optical flow measurement system, which are determined by various factors, such as the resolution, the image acquisition rate, and the focal length of the camera. 
It can be seen from the above equations that the position of the MAV in the $X$- and $Y$-axes are not observable from the lateral flows, they still can be used in stabilizing the position by maintaining zero lateral flows or controlling the velocity with an appropriate constant. 

When the MAV approaches the surface, the optical flow captured from the camera shows a divergent pattern. It corresponds to the flow divergence $\vartheta_z$, which represents the velocity of the features in the image plane along the line of sight and can be formulated as 
\begin{equation}
\label{eq:flowdivergenceZ}
\vartheta_z = c_z \frac{v_z-v_z^r}{d_z-d_z^r},
\end{equation}
where $c_z$ is an unknown positive constant. 
When the MAV is approaching the platform, $v_z-v_z^r$ is negative, $d_z-d_z^r$ is always positive, and thus, $\vartheta_z$ will be negative. The constant flow divergence approach \cite{zhou2021extended} has been used to control the vertical dynamics of MAVs by keeping the flow divergence a negative constant for landing, i.e., $\vartheta_z=\vartheta_z^*$. As a result, the time response of the distance $d_z-d_z^r$ will decrease exponentially to zero.

\section{Data-Driven Incremental nonlinear dynamic inversion}
\label{sec:INDI}
As depicted in the previous section, the system dynamics in Eq. (\ref{eq:positionDynamicsExp}) is input non-affine, the model parameters are unknown, and the full states are not directly measurable. Therefore, this paper proposes to use a data-driven INDI method to control the MAV to track and land on a moving platform. 

\subsection{Reformulation of the optical flow tracking control problems}

The INDI method can be derived from the system as in Eqs. (\ref{eq:positionDynamics1}, \ref{eq:positionDynamicsExp}-\ref{eq:flowdivergenceZ}) or in a more general form for nonlinear systems:
\begin{align}
\label{eq:nonlinearSystem}
\dot{\bm{x}} &= \bm{f}(\bm{x},\bm{u}),\\
\label{eq:nonlinearOutput}
\bm{y} &= \bm{h}(\bm{x}),
\end{align}
where $\bm{x} \in \mathcal{R}^n$ is the system state, $\bm{u} \in \mathcal{R}^m$ is the system input, $\bm{y} \in \mathcal{R}^p$ is the system output or observations, $\bm{f} : \mathcal{R}^{n+m} \rightarrow \mathcal{R}^n$ and $\textbf{\textit{h}} : \mathcal{R}^n \rightarrow \mathcal{R}^p$ are smooth vector fields. If $p<m$ or $p>m$, it becomes an over-determined or under-determined control problem, and control allocation techniques or weighted least square methods can be used to tackle these problems, respectively.  The rest of the paper assumes that the numbers of inputs and outputs are equal, $p=m$, for both theoretical derivations and practical applications. 

In optical flow tracking problems, the observation involves the dynamics of the system and the platform as indicated in Eqs. (\ref{eq:opticalFlowX}-\ref{eq:flowdivergenceZ}). Thus, the state space needs to be augmented to incorporate both the states of the system and the referred platform, i.e., $[\bm{d}^T, \bm{v}^T, {\bm{d}^r}^T, {\bm{v}^r}^T]^T$. 
In biological systems, relative measurements are often utilized in navigation and control. For instance, many animals rely on relative motion cues in their visual system to track moving targets and navigate their environment. 
Similarly, the formulation of this control task focuses on the relative position and velocity $[(\bm{d}-\bm{d}^r)^T, (\bm{v}-\bm{v}^r)^T]^T$, which reduces the dimensionality of the problem. 
In addition, the relative position in the $X$- and $Y$-axes are not observable as mentioned in the previous section. The state of interest can be further reduced and defined as $\bm{x}=[\tilde{d}_z, \tilde{v}_x, \tilde{v}_y, \tilde{v}_z]^T=[d_z-d_z^r, v_x-v_x^r, v_y-v_y^r, v_z-v_z^r]^T$. 
Thus, the dynamics of the relative position can be reformulated as follows:
\begin{align}
\label{eq:relativePositionDynamics1}
\dot{\tilde{d}}_z &= \tilde{v}_z,\\
\label{eq:relativePositionDynamics2}
\dot{\tilde{v}}_x &=  \frac{1}{M} \cos \phi \sin \theta T 
+ \frac{1}{M} \delta_x - \dot{v}_x^r ,\\
\label{eq:relativePositionDynamics3}
\dot{\tilde{v}}_y &=  -\frac{1}{M} \sin \phi T 
+ \frac{1}{M} \delta_y - \dot{v}_y^r ,\\
\label{eq:relativePositionDynamics4}
\dot{\tilde{v}}_z &= \frac{1}{M} \cos \phi \cos \theta T -g 
+ \frac{1}{M} \delta_z - \dot{v}_z^r .
\end{align}

\subsection{Incremental nonlinear dynamic inversion}

Since the observation $\bm{y}$ is nonlinear in the state $\bm{x}$, the first step is to find the input-output relation by differentiating the output until the input appears explicitly:
\begin{align}
\label{eq:opticalXDiff}
\dot{\vartheta}_x & \begin{aligned}[t]
&= c_x \left[ \frac{\dot{\tilde{v}}_x}{\tilde{d}_z} - \frac{\tilde{v}_x \tilde{v}_z}{\tilde{d}_z^2} \right]\\
&= c_x \left[ \frac{(\cos \phi \sin \theta T   +\delta_x  - M\dot{v}_x^r)}{M \tilde{d}_z} - \vartheta_x \vartheta_z \right],
\end{aligned}\\
\label{eq:opticalYDiff}
\dot{\vartheta}_y & \begin{aligned}[t]
&= c_y \left[ \frac{\dot{\tilde{v}}_y}{\tilde{d}_z} - \frac{\tilde{v}_y \tilde{v}_z}{\tilde{d}_z^2} \right]\\
&= c_y \left[ \frac{(- \sin \phi  T   +\delta_y  - M\dot{v}_y^r)}{M \tilde{d}_z} - \vartheta_y \vartheta_z \right],
\end{aligned}\\
\label{eq:flowdivergenceZDiff}
\dot{\vartheta}_z & \begin{aligned}[t]
&=c_z \left[ \frac{\dot{\tilde{v}}_z}{\tilde{d}_z} - \frac{\tilde{v}_z^2}{\tilde{d}_z^2} \right]\\
&= c_z \left[ \frac{(\cos \phi \cos \theta T  -Mg +\delta_z  - M\dot{v}_z^r)}{M \tilde{d}_z} - \vartheta_z^2 \right].
\end{aligned}
\end{align}
The above input-output relation can be written as
\begin{equation}
\label{eq:MIMOinputNonAffineGeneral}
\dot{\bm{y}} 
= \bm{\alpha}(\bm{x},\bm{u},\bm{\delta},\dot{\bm{v}}^r),
\end{equation}
which is nonlinear and nonaffine in the input. 
Now, taking the first-order Taylor series expansion results in:
\begin{equation}
\label{eq:taylorIO}
\begin{split}
\dot{\bm{y}}
\approx ~  &\dot{\bm{y}}_0
+ \frac{\partial \bm{\alpha}(\bm{x},\bm{u},\bm{\delta},\dot{\bm{v}}^r) }{\partial \bm{x}} |_{\bm{x}_0, \bm{u}_0,\bm{\delta}_0,\dot{\bm{v}}^r_0} (\bm{x}-\bm{x}_0) \\
&+ \frac{\partial \bm{\alpha}(\bm{x},\bm{u},\bm{\delta},\dot{\bm{v}}^r) }{\partial \bm{u}} |_{\bm{x}_0, \bm{u}_0,\bm{\delta}_0,\dot{\bm{v}}^r_0} (\bm{u}-\bm{u}_0) \\
&+ \frac{\partial \bm{\alpha}(\bm{x},\bm{u},\bm{\delta},\dot{\bm{v}}^r) }{\partial \bm{\delta}} |_{\bm{x}_0, \bm{u}_0,\bm{\delta}_0,\dot{\bm{v}}^r_0} (\bm{\delta}-\bm{\delta}_0) \\
&+ \frac{\partial \bm{\alpha}(\bm{x},\bm{u},\bm{\delta},\dot{\bm{v}}^r) }{\partial \dot{\bm{v}}^r } |_{\bm{x}_0, \bm{u}_0,\bm{\delta}_0,\dot{\bm{v}}^r_0} (\dot{\bm{v}}^r-\dot{\bm{v}}^r_0),
\end{split}
\end{equation}
where the second or higher-order terms have been neglected in this equation.  

This equation can be further simplified with a time-scale separation assumption. This assumption holds when the change in the control input is considered significantly faster than the change in the system state with high sampling frequency and instantaneous control effects \cite{van2018stability}. 
Note that in optical flow control problems, this assumption holds only when $1/\tilde{d}_z$ is bounded \cite{zhou2021extended}. 
This assumption is put into practice by setting a threshold for the height $\tilde{d}_z$, 
below which the system will shut down. 
In addition, it is assumed that the aerodynamic forces $\bm{\delta}$ and platform dynamics are slower than the system dynamics. This allows the controller to focus on identifying the direct input-output relationship and generate the control input more efficiently. 
With these assumptions, Eq. (\ref{eq:taylorIO}) can be further simplified to
\begin{equation}
\label{eq:taylorIOTS}
\dot{\bm{y}} 
\approx  \dot{\bm{y}}_0
+ G(\bm{x}_0, \bm{u}_0) (\bm{u}-\bm{u}_0) ,
\end{equation}
where  $\bm{u}=[\theta , \phi, T]^T$ and
\begin{equation}
\label{eq:linearG}
\begin{split}
&G(\bm{x}, \bm{u})  = 
\frac{1}{M \tilde{d}_z} \cdot \\
& \begin{bmatrix}
c_x \cos \phi \cos \theta T &  -c_x \sin \phi \sin \theta T &   c_x \cos \phi \sin \theta\\
0 &  -c_y\cos \phi T &  -c_y\sin \phi \\
- c_z\cos \phi \sin \theta T &  -c_z\sin \phi \cos \theta T & c_z\cos \phi \cos \theta 
\end{bmatrix}.
\end{split}
\end{equation}

To solve the above nonlinear control problem, the virtual control input is introduced to replace the differentiated output $\dot{\bm{y}} =\bm{\nu}$, and the control increment can be found by inverting Eq. (\ref{eq:taylorIOTS}):
\begin{equation}
\label{eq:INDIcontrolOutput}
\bm{u}-\bm{u}_0
=  G(\bm{x}_0, \bm{u}_0)^{-1} (\bm{\nu} - \dot{\bm{y}}_0).
\end{equation}
This matrix $G$ is a nonlinear function of Euler angles, $\phi$ and $\theta$, the total thrust $T$, and unknown parameters $c_x$, $c_y$, and $c_z$. It also varies nonlinearly with the distance above the platform $\tilde{d}_z$, which is not directly measurable with the optical flow sensors. Thus, it is very difficult to measure all the relevant variables and calculate matrix $G$ from Eq. (\ref{eq:linearG}) at every time step. 
Another option to find the control increment is to identify $G$ online using least square techniques from input and output data with the time sequence, which can be rewritten as  
\begin{equation}
\label{eq:taylorIOTSdiscrete}
\dot{\bm{y}}_k -
 \dot{\bm{y}}_{k-1}
\approx  G_{k-1} \cdot (\bm{u}_{k}-\bm{u}_{k-1}) .
\end{equation}
In this case, the explicit representation of the time-varying matrix $G$ in Eq. (\ref{eq:linearG}) is unnecessary, but it needs to be identified and inverted at every time step. 

\subsection{Inverted G matrix identification}
To reduce the computational load of this data-driven approach, this paper proposes to directly identify the inverted $G$ matrix, which is denoted as $G^\dagger=G^{-1}$, from input and output data with time sequence $t_{k}, t_{k-1}, t_{k-2}, \cdots$ as follows:
\begin{equation}
\label{eq:taylorIOTSinverse}
\bm{u}_{k}-\bm{u}_{k-1}
\approx  G^\dagger_{k-1} \cdot (\dot{\bm{y}}_{k} -
 \dot{\bm{y}}_{k-1}) .
\end{equation}
The parameters in matrix $G^\dagger \in \mathcal{R}^{m \times p}$ can be identified using the Recursive Least Square (RLS) approach. 
The control command can be calculated directly with the identified matrix $\widehat{G}^\dagger$:
\begin{equation}
\begin{split}
\label{eq:INDIcontrolOutput}
\bm{u}=\bm{u}_k
+ \widehat{G}^\dagger_{k-1} \cdot (\bm{\nu}_k - \dot{\bm{y}}_k).
\end{split}
\end{equation}
The virtual input $\bm{\nu}_{k}$ at the current time $t_k$ can be designed as a feedback linear control with constant proportional, integral, and derivative feedback gains: 
\begin{equation}
\label{eq:EINDIvirtualFD}
\bm{\nu}_{k} = -K_p(\bm{y}_k-\bm{\vartheta}^*) - K_i \int^t_0 (\bm{y}_k-\bm{\vartheta}^*)dt - K_d ~\dot{\bm{y}}_k,
\end{equation}
where $\bm{\vartheta}^*$ is a constant vector of the desired optical flow. In the tracking platform task, we will have $\bm{\vartheta}^*=\bm{0}$, and in the landing on a moving platform task, we will have $\bm{\vartheta}^*=[0~0~\vartheta_z^*]^T$. The desired flow divergence $\vartheta_z^*$ for landing is a negative constant, and by following the constant flow divergence, the MAV will have a soft land on the platform with the time response \cite{zhou2021extended}:
\begin{equation}
\label{eq:heightperfectlanding}
\tilde{d}_z(t) = \tilde{d}_{z_0}e^{\frac{\vartheta_z^*}{c_z}t},
\end{equation}
where $\tilde{d}_{z_0}$ is the initial height above the landing surface.

In the data-based INDI methods, where the matrix $G$ is unknown and time-varying, the control command can be calculated by first identifying $G$ and then calculating its inversion in the traditional way, or directly identifying the inverse matrix $G^\dagger$ using our proposed algorithm. 
The direct identification of $G^\dagger$ has a lower time complexity and computation load compared to the traditional method. This is because the inversion step, which requires  $\mathcal{O}(p^3)$ computational effort, is no longer necessary. 
Therefore, the direct identification of $G^\dagger$ results in a more efficient INDI control implementation.

The proposed method of directly identifying the inverse matrix $G^\dagger$ also has advantages in terms of accuracy and stability, especially for optical flow control. The matrix $G$ increases to very large values and changes very quickly when the vehicle is approaching the platform, as it is proportional to  $1/\tilde{d}_z$. This can result in difficulties for the control system to accurately estimate the matrix $G$. On the other hand, the inverse matrix $G^\dagger$ tends to converge as the vehicle approaches the platform, which makes the updating process more straightforward and precise. 
Furthermore, by eliminating the inversion step, the proposed method reduces the chances of numerical errors and improves the stability of the control system, especially when dealing with the possibility of ill-conditioned matrices. 
In conclusion, the direct identification of the inverse matrix $G^\dagger$ offers both computational efficiency and improved accuracy and stability of the data-driven INDI approach.


\section{Numerical Simulations}


In this section, the feasibility of the proposed optical flow control method will be demonstrated through three-dimensional tracking control simulations. 
This will be achieved by comparing the INDI control with both 1) the conventional method  that identifies the $G$ matrix and calculates its inverse at every time step and 2) the proposed method that directly identifies the inverse matrix $G^\dagger$. 
The simulations are performed with an MAV having a mass of $M=1.2~kg$, proportionality constants $c_x, c_y, c_z$ equal to $1$, and an initial height of $3~m$. 
For a fair comparison, the initial parameters and the INDI outer loop implementations are the same. 
The outer loop virtual input is designed using a linear feedback control with a constant proportional gain $K_p=1$. 
All these simulations are conducted in the MATLAB environment on a computer with 1.8 GHz CPU and 40 GB of RAM.

\begin{figure}[!t]
	\centering
	\input{INDI_G.tex}
	\caption{The MAV lands on a moving platform with INDI control identifying and inverting the $G$ matrix at each step. Each column presents the positions of the MAV $\bm{d}$ (the dashed line represents the position of the platform $\bm{d}^r$), optical flows $\bm{\vartheta}$ (the dashed line represents the tracked setpoint $\bm{\vartheta}^*=[0,0,-0.1]$), control inputs $\bm{u}=[\theta, \phi, T]^T$, and diagonal elements of the identified matrix $\hat{G}$ (the solid line represents the true value) in the (a) $X$-axis, (b) $Y$-axis, and (c) $Z$-axis.}
	\label{fig:INDI_G}
\end{figure}
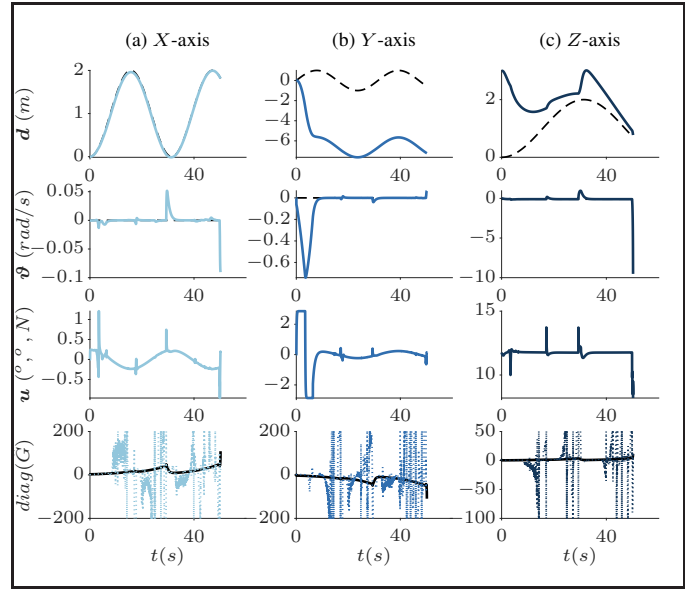

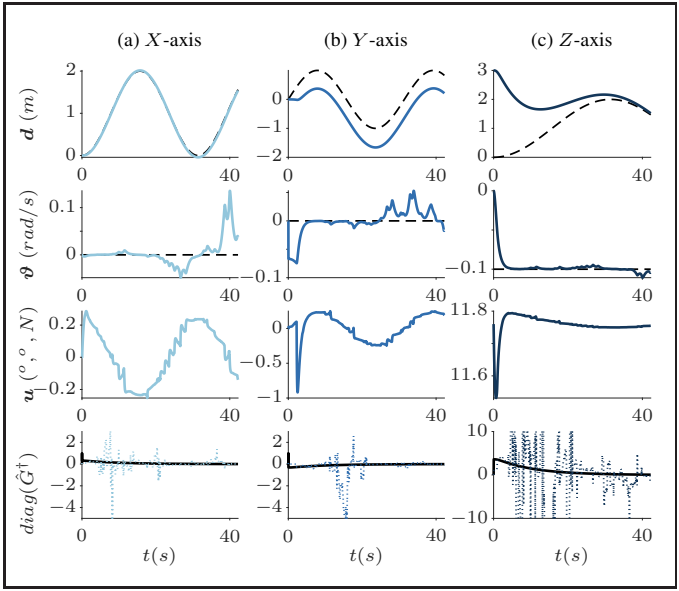
\begin{figure}[!t]
	\centering
	\input{INDI_Ginv.tex}
	\caption{The MAV lands on a moving platform with INDI control identifying the inverse matrix $G^\dagger$ directly. Each column presents the positions of the MAV $\bm{d}$ (the dashed line represents the position of the platform $\bm{d}^r$), optical flows $\bm{\vartheta}$ (the dashed line represents the tracked setpoint $\bm{\vartheta}^*=[0,0,-0.1]$), control inputs $\bm{u}=[\theta, \phi, T]^T$, and diagonal elements of the inverse matrix $\hat{G}^\dagger$ (the solid line represents the true value) in the (a) $X$-axis, (b) $Y$-axis, and (c) $Z$-axis.}
	\label{fig:INDI_Ginv}
\end{figure}

Figure \ref{fig:INDI_G} presents the performance of the INDI control for an MAV landing on a moving platform with a constant flow divergence setpoint $\vartheta^*_z=-0.1~rad/s$. 
The reference platform undergoes a three-dimensional translational movement in a sinusoidal form as shown by the dashed line in the first row. 
In this simulation, the matrix $G$ is identified using the RLS technique, and its inverse is calculated at every time step to generate the control inputs according to Eq. (\ref{eq:INDIcontrolOutput}). 
As shown in Eq. (\ref{eq:linearG}), the non-diagonal elements of the matrix $G$ are small because the pitch angle $\theta$ and roll angle $\phi$ are small, thus the figures only depict the diagonal elements of the identified matrix $\hat{G}$. The true values of the matrix $G$ diagonal elements, which correspond to the motion in the $X$-, $Y$-, and $Z$-axes, are calculated from Eq. (\ref{eq:linearG}) and presented in solid lines. 

It can be seen that, as the MAV approaches the platform, the observed optical flow is highly sensitive to even minor input variations, and the diagonal elements of the matrix $G$ increase rapidly. In this case, it is essential to update the estimated $G$ matrix at a sufficient rate to adapt the control law to ensure a smooth landing. Failure to do so will result in degraded control performance. 
Therefore, in this optical flow control task, especially when the height $\tilde{d}_z$ is small, the conventional method  of identifying and inverting the matrix $G$ is sensitive to the choice of the forgetting factor $\gamma$ in the RLS approach. 
To facilitate rapid parameter updates, the simulation employs a value of $\gamma=0.8$. This choice, however, may lead to oscillations in the estimation. Conversely, large values of $\gamma$ will impede the update rate of matrix $G$, resulting in even worse performance. 
Another practical approach is to set a higher threshold for the height $\tilde{d}_z$ and shut down the system when it falls below the threshold. This approach, however, requires additional sensing and operations. 

On the other hand, the inverse matrix $G^\dagger$, which is proportional to the height  $\tilde{d}_z$, converges as the MAV approaches the landing platform as depicted in Figure \ref{fig:INDI_Ginv}. 
When the height $\tilde{d}_z$ is small, the directly identified $G^\dagger$ decreases correspondingly to small values and generates small control increments to facilitate a smooth landing of the MAV. 
In this scenario, the forgetting factor $\gamma$ may be increased to achieve a more stable parameter update than what is illustrated in the figure. 
Although both the matrix $G$ and its inverse matrix $G^\dagger$  is time-varying, the direct identification of $G^\dagger$ will be less sensitive to the choice of the forgetting factor in the RLS identification process and the choice of the threshold for the height $\tilde{d}_z$. 
Furthermore, the computational efficiency of the proposed INDI control is demonstrated by a decrease in the CPU time required to run a 50-second-simulation from $0.1062~s$ in the traditional INDI control to $0.0681~s$ in the proposed INDI control (averaged over 20 simulations).


\section{Flight Tests}
This section presents the experimental validation of the proposed INDI control scheme implemented on an MAV. Several flight tests were performed in an indoor environment under various challenging conditions: 1) landings with different control parameters on a static and flat landing surface, and 2) hovering and landings over moving and undulating landing surfaces. The experimental setup and the results of all the flight tests are presented and discussed in this section.


\subsection{Flight Test Setup}
\label{subsec:Setup}
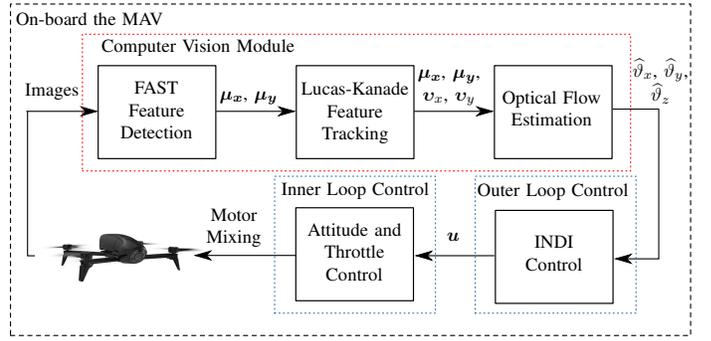
\begin{figure}[!t]
	\centering
	\input{HardwareSoftware}
	\caption{Flight test setup using an MAV (Parrot Bebop 2) with an open source autopilot (Paparazzi). All the computer vision and control algorithms run onboard the MAV. In the vision module, FAST detects corners ($\bm{\mu_x}$, $\bm{\mu_y}$) in each image captured by the downward camera. Then, Lucas-Kanade tracks these corners in the next image and computes the flow vectors ($\bm{\upsilon_x}$, $\bm{\upsilon_y}$). Lastly, optical flow in the $X^b$-, $Y^b$-, $Z^b$-axes ($\bm{\widehat{\vartheta}}$) as defined in this paper are estimated. In the outer loop control module, the INDI controller receives $\bm{\widehat{\vartheta}}$ as the input and commands the attitude and throttle of the MAV ($\bm{u}$).}
	\label{fig:HardwareSoftware}
\end{figure}

The experimental setup uses a Parrot Bebop 2 MAV with the Paparazzi autopilot software\footnote{\label{pprz}Paparazzi Autopilot:~\url{http://wiki.paparazziuav.org}}, as illustrated in Figure \ref{fig:HardwareSoftware}. It is equipped with standard sensors, including an inertia measurement unit, a sonar sensor, and two cameras (front and bottom). The bottom camera (facing downward) is of particular interest for hovering and landing purposes. The sonar sensor was enabled for logging height during the tests to evaluate the landing performance.

In the autopilot software, two modules were specifically created for the flight tests: the computer vision and the outer loop control, as depicted in Figure~\ref{fig:HardwareSoftware}. The computer vision module processes the images captured from the camera during the flight tests and implements computer vision algorithms, including the Features from Accelerated Segment Test (FAST) algorithm \cite{rosten2006machine} for feature detection and the Lucas-Kanade tracker \cite{bouquet2000pyramidal} for feature tracking. 

By taking the average of the flow vectors in the $X^b$- and $Y^b$-axes ($\bm{\upsilon}_x$,~$\bm{\upsilon}_y$) obtained from the feature tracking, optical flow $\widehat{\vartheta}_x$ and $\widehat{\vartheta}_y$ can be computed. To estimate flow divergence $\widehat{\vartheta}_z$, the approach in \cite{ho2018adaptive} utilizing Eq. (\ref{equation:size_divergence}) is employed. This involves computing the image distances of every two features between consecutive images $\kappa$, and then computing the average ratio of the change in the image distances to its previous image distance with a known time interval $\Delta t$, resulting in the estimation of flow divergence $\widehat{\vartheta}_z$ as follows:

\begin{equation}
\widehat{\vartheta}_z = \frac{1}{M}\cdot\frac{1}{\Delta t}\sum_{i=1}^{M}[\frac{\kappa_{(t-\Delta t), i}-\kappa_{t, i}}{\kappa_{(t-\Delta t), i}}],
\label{equation:size_divergence}
\end{equation}
\noindent where $M$ is the total number of tracked corners. A low-pass filter was introduced to minimize the estimation noise. 

The outer loop control module, which implements the INDI control scheme detailed in Section \ref{sec:INDI}, receives output from the computer vision module and controls the attitudes and throttle of the MAV. The flight tests were started with manual take-off to a specified position and then switched to automatic mode, where only the required modules executed the designated tests. The height measurement from the sonar sensor was recorded for performance validation and analysis purposes, but not utilized in the control process.

\subsection{Static and flat landing surface}
\label{subsec:static}
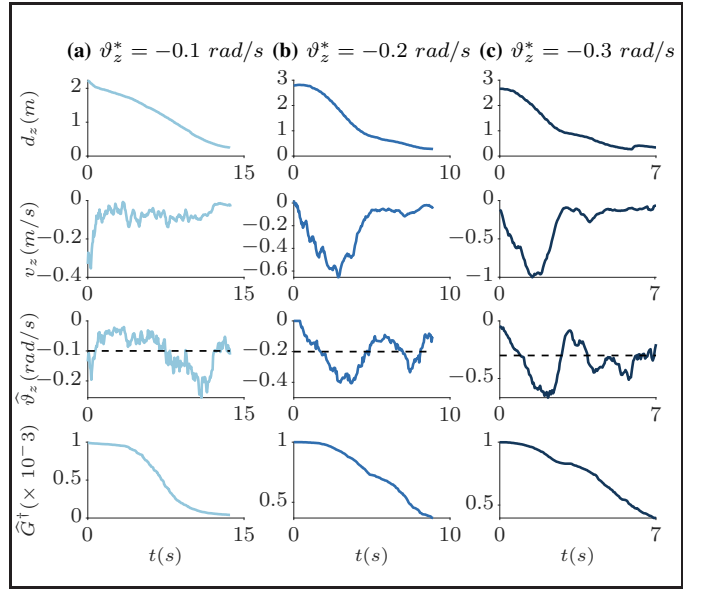
\begin{figure}[!t]
	\centering
	\input{landing_static_setpoints_inverseG.tex}
	\caption{MAV Landings on a static and flat ground by tracking different flow divergence setpoints: (a) $\vartheta_z^*=-0.1~rad/s$, (b) $\vartheta_z^*=-0.2~rad/s$, (c) $\vartheta_z^*=-0.3~rad/s$. Each column presents the MAV height $d_z$, vertical velocity $v_z$, flow divergence $\widehat{\vartheta}_z$, and inverse matrix $\widehat{G}^\dagger$ for the respective flight test. All the tracked setpoints ($\vartheta_z^*$) are plotted as dashed lines.}
	\label{fig:landing_static_setpoints}
\end{figure}

Figure \ref{fig:landing_static_setpoints} presents multiple landing tests of the MAV on a static and flat landing surface, in which three different setpoints of flow divergence, i.e., $-0.1~rad/s$, $-0.2~rad/s$, and $-0.3~rad/s$, are tracked using the proposed INDI control. As the setpoint value increases, the landing velocity also increases, making the tracking task more demanding for the controller as it must efficiently accommodate the rapidly changing velocity of the MAV. 

The results depicted in Figure \ref{fig:landing_static_setpoints} clearly demonstrate the ability of the controller to accurately track the designated setpoints. It also exhibits a quick response to even larger setpoints, indicating its robustness and efficiency. As a result, for all landings, both height and vertical velocity of the MAV decay exponentially to zero at almost the same time, as desired. Since this control implementation is designed specifically for vertical landings, the inverse matrix $\widehat{G}^\dagger$ contains only a single element. It is noteworthy that the identified $\widehat{G}^\dagger$ exhibits an exponential decay and convergence, analogous to the change in height. This observation supports the validity of the theoretical derivation of $G^\dagger$, which is a function of height.

\subsection{Moving and undulating landing surface}
\label{subsec:moving}
\begin{figure*}[!t]
	\centering
	\input{moving_platform4.tex}
	\caption{MAV Hovering and Landings on a moving and undulating ground. The columns (a) and (b) depict the hovering (by tracking $\vartheta_z^*=0~rad/s$, $\vartheta_y^*=0~rad/s$) and landing (by tracking $\vartheta_z^*=-0.2~rad/s$, $\vartheta_y^*=0~rad/s$) results of the MAV, respectively, above a moving and flat surface, while the column (c) displays the landing (by tracking $\vartheta_z^*=-0.1~rad/s$) results of the MAV during translation (by tracking $\vartheta_y^*=-0.8~rad/s$) over an undulating ground. Each column shows the stacked images of the flight scene, MAV height $d_z$, flow divergence $\widehat{\vartheta}_z$, and optical flow in  the $Y^b$-axis $\widehat{\vartheta}_y$ for the respective flight test. All the tracked setpoints ($\vartheta_z^*$, $\vartheta_y^*$) are plotted as dashed lines.}
	\label{fig:moving_platform}
\end{figure*}
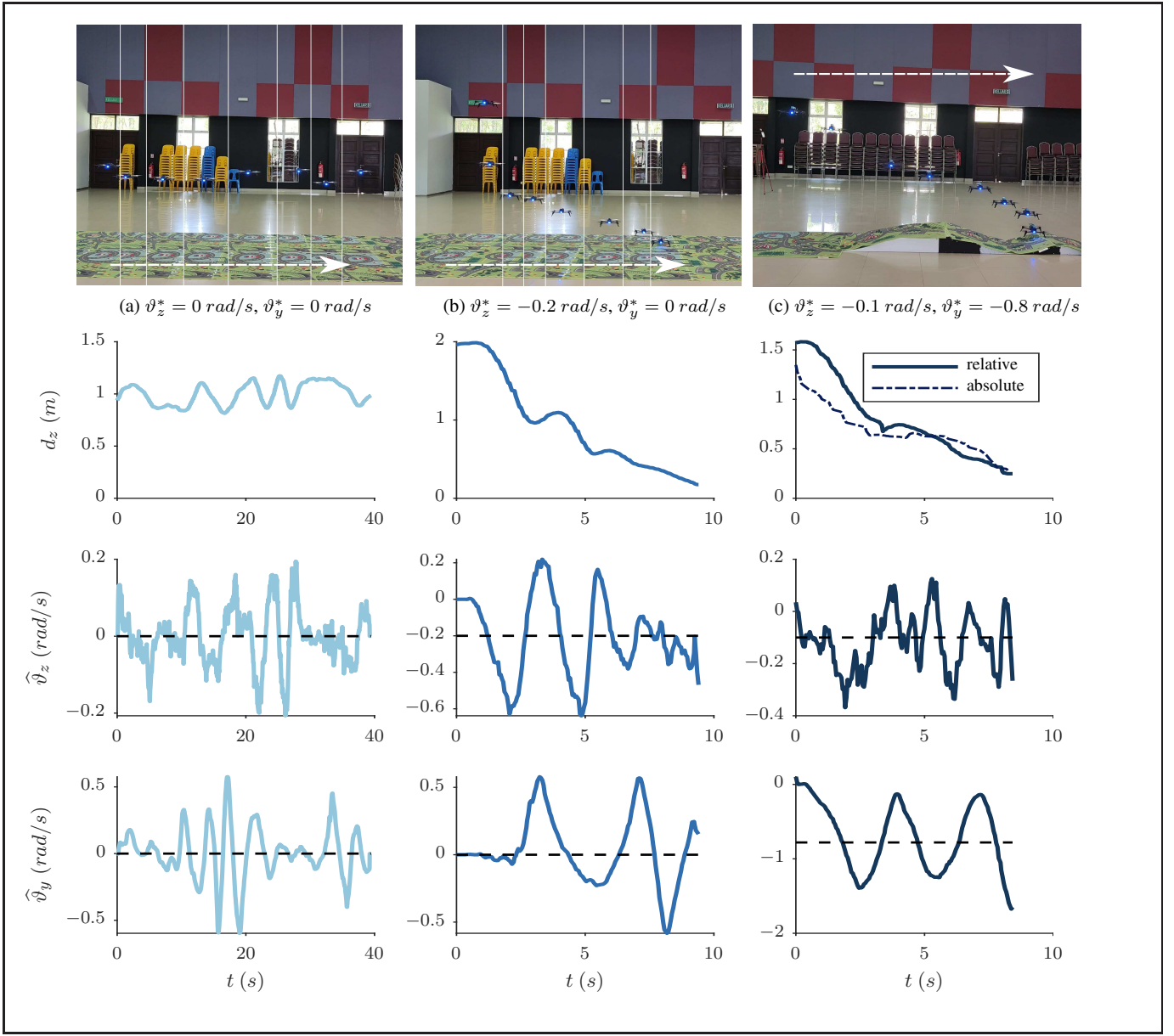

The results of the hovering and landing tests conducted on a dynamically changing landing surface, characterized by wave-like fluctuations, are depicted in Figure \ref{fig:moving_platform}. These flight tests were carried out to verify the efficiency and robustness of the proposed INDI control scheme in the presence of lateral and vertical movements of the landing surface. 

The first test, whose results are presented in Figure \ref{fig:moving_platform} (a), involved hovering of the MAV over a surface undergoing lateral movement by utilizing the controller to regulate the flow divergence ($\vartheta_z^*=0~rad/s$) and optical flow in the $Y^b$-axis ($\vartheta_y^*=0~rad/s$). The results in this figure show that both setpoints are well tracked using the INDI controller even for a laterally moving surface. This can be observed from the nearly constant height and unchanged position of the MAV with respect to the surface feature shown in the stacked image.

Besides the regulation task, Figure \ref{fig:moving_platform} (b) depicts the results of a flight test on a surface undergoing lateral movement during landing, rather than hovering. The controller was tasked with tracking a constant flow divergence setpoint, i.e., $\vartheta_z^*=-0.2~rad/s$ (a median value of the setpoints was selected from the tests performed in Section \ref{subsec:static}) for landing and maintaining the regulation of the optical flow in the $Y^b$-axis ($\vartheta_y^*=0~rad/s$) for following the moving surface. The results indicate that the controller was adapting the changes at the beginning, as the estimated $\widehat{\vartheta}_z$ shows a slower response causing a small fluctuation of height at $t\approx4~s$. Eventually, the controller is able to accurately track the setpoint $\vartheta_z^*$ and land the MAV with an exponential decay of the height $d_z$. 

In order to evaluate the capabilities of the controller under more challenging conditions, the last flight test stimulated the vertical movement of the landing surface by translating the MAV over an undulating terrain. The lateral movement of the MAV was produced by tracking a setpoint of optical flow in the $Y^b$-axis ($\vartheta_y^*=-0.8~rad/s$). To demonstrate that the MAV can safely land on the uneven terrain, a flow divergence setpoint $\vartheta_z^*=-0.1~rad/s$ was chosen. Since the sonar measurement only provides the height relative to the landing surface, an ultra-wideband (UWB) tag was attached to the MAV together with four UWB anchors located at four corners of the test area to measure the absolute height of the MAV using Double-Sided Two-Way Ranging \cite{ching2022ultra}. The absolute height allows us to analyze at which instance the MAV changing its velocity or even stopping. The results show that the controller adapted well to the translation over an undulating terrain as the tracking of the flow divergence $\vartheta_z^*$ is accurate leading to the exponential decay of the relative height, as desired. Additionally, the result from the absolute height shows that the MAV was descending in a slow pace when it encountered the first peak ($t\approx4~s$). This is due to the fact that the MAV detected its movement closer to the landing surface at that instance. After it moved over the first peak, the MAV continued to descend and land safely on the next peak of the undulating surface while translating.




\section{Conclusion} 
\label{sec:conclusion}

This paper formulated a novel data-driven Incremental Nonlinear Dynamic Inversion (INDI) method to address nonlinear optical flow control problems with rapidly changing and height-dependent system output dynamics. 
The control effectiveness matrix $G$ in INDI for optical flow control was found to be time-varying with the distance above the platform, which is not directly measurable. 
Consequently, an online identification of $G$ from input and output data, followed by an inversion, at every time step was deemed necessary to calculate the control increments. To this end, the direct identification of the inverse matrix $G\dagger$ was proposed in order to eliminate the inversion step, thus reducing the chances of numerical errors and enhancing the accuracy and stability of the data-driven INDI approach. 
The proposed method was applied and compared with the conventional method in numerical simulations for controlling an MAV during landing on a moving platform. It showed an improved control performance and also computational efficiency. 
The proposed method was also validated through real-world flight tests, successfully demonstrating the ability to track and land on various surfaces with desired height and velocity decay. 
Future work will focus on 1) extending the incremental model to deal with measurement noises and delays, scaleless system outputs, or fast dynamics, and 2) integrating this efficient optical flow control method with a guidance strategy from other visual cues for precise navigation of small flying robots in obstacle-dense environments. 


\section*{Acknowledgments}
This study is sponsored by Malaysian Ministry of Higher Education with the Fundamental Research Grant Scheme (FRGS) [grant number FRGS/1/2020/TK0/USM/03/11].

\bibliographystyle{unsrtnat}
\bibliography{references}

\end{document}

%% file: intro.tex
\begin{psfrags}%
\psfragscanon%
\newcommand{\tsize}{1}
\newcommand{\tsizeb}{0.75}
%
\psfrag{a}[b][b][\tsizeb]{\color[rgb]{1,1,1}\setlength{\tabcolsep}{0pt}\begin{tabular}{c}$\vartheta_z^*=-\xi_1$, $\vartheta_y^*=-\xi_2$\end{tabular}}%
\psfrag{b}[b][b][\tsizeb]{\color[rgb]{1,1,1}\setlength{\tabcolsep}{0pt}\begin{tabular}{c}$\tilde{d}_z=\tilde{d}_{z_0}e^{-\xi_1 t}$\\$\tilde{v}_z=-\xi_1 \tilde{d}_{z_0}e^{-\xi_1 t}$\end{tabular}}%

\includegraphics[width=0.45\textwidth]{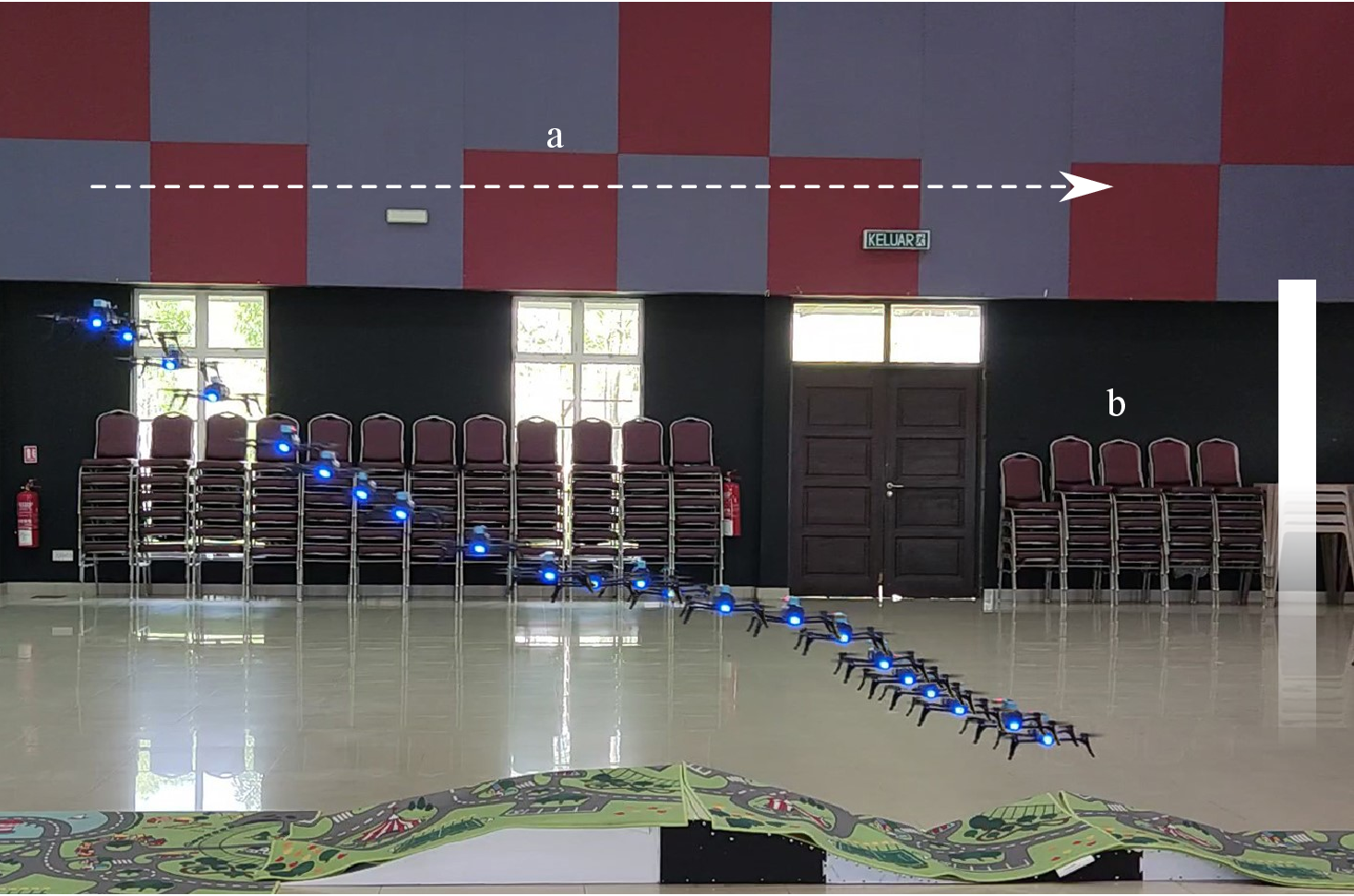}
\end{psfrags}%
%

%% file: uav_body_axis.tex
\begin{psfrags}%
\psfragscanon%
\newcommand{\tsize}{1}
\newcommand{\tsizeb}{0.75}
%
\psfrag{x}[b][b][\tsizeb]{\color[rgb]{0,0,0}\setlength{\tabcolsep}{0pt}\begin{tabular}{c}$X^b$\end{tabular}}%
\psfrag{y}[b][b][\tsizeb]{\color[rgb]{0,0,0}\setlength{\tabcolsep}{0pt}\begin{tabular}{c}$Y^b$\end{tabular}}%
\psfrag{z}[b][b][\tsizeb]{\color[rgb]{0,0,0}\setlength{\tabcolsep}{0pt}\begin{tabular}{c}$Z^b$\end{tabular}}%
\psfrag{d}[l][c][\tsizeb]{\color[rgb]{0,0,0}\setlength{\tabcolsep}{0pt}\begin{tabular}{c}$Y^I$\end{tabular}}%
\psfrag{e}[b][b][\tsizeb]{\color[rgb]{0,0,0}\setlength{\tabcolsep}{0pt}\begin{tabular}{c}$X^I$\end{tabular}}%
\psfrag{f}[b][b][\tsizeb]{\color[rgb]{0,0,0}\setlength{\tabcolsep}{0pt}\begin{tabular}{c}$Z^I$\end{tabular}}%

\psfrag{o}[t][t][\tsizeb]{\color[rgb]{0,0,0}\setlength{\tabcolsep}{0pt}\begin{tabular}{c}$O^b$\end{tabular}}%
\psfrag{p}[t][t][\tsizeb]{\color[rgb]{0,0,0}\setlength{\tabcolsep}{0pt}\begin{tabular}{c}$O^I$\end{tabular}}%

\psfrag{b}[l][b][\tsize]{\color[rgb]{0,0,0}\setlength{\tabcolsep}{0pt}\begin{tabular}{c}$b$: body\end{tabular}}%
\psfrag{w}[l][b][\tsize]{\color[rgb]{0,0,0}\setlength{\tabcolsep}{0pt}\begin{tabular}{c}$I$: inertial\end{tabular}}%
\includegraphics[width=0.35\textwidth]{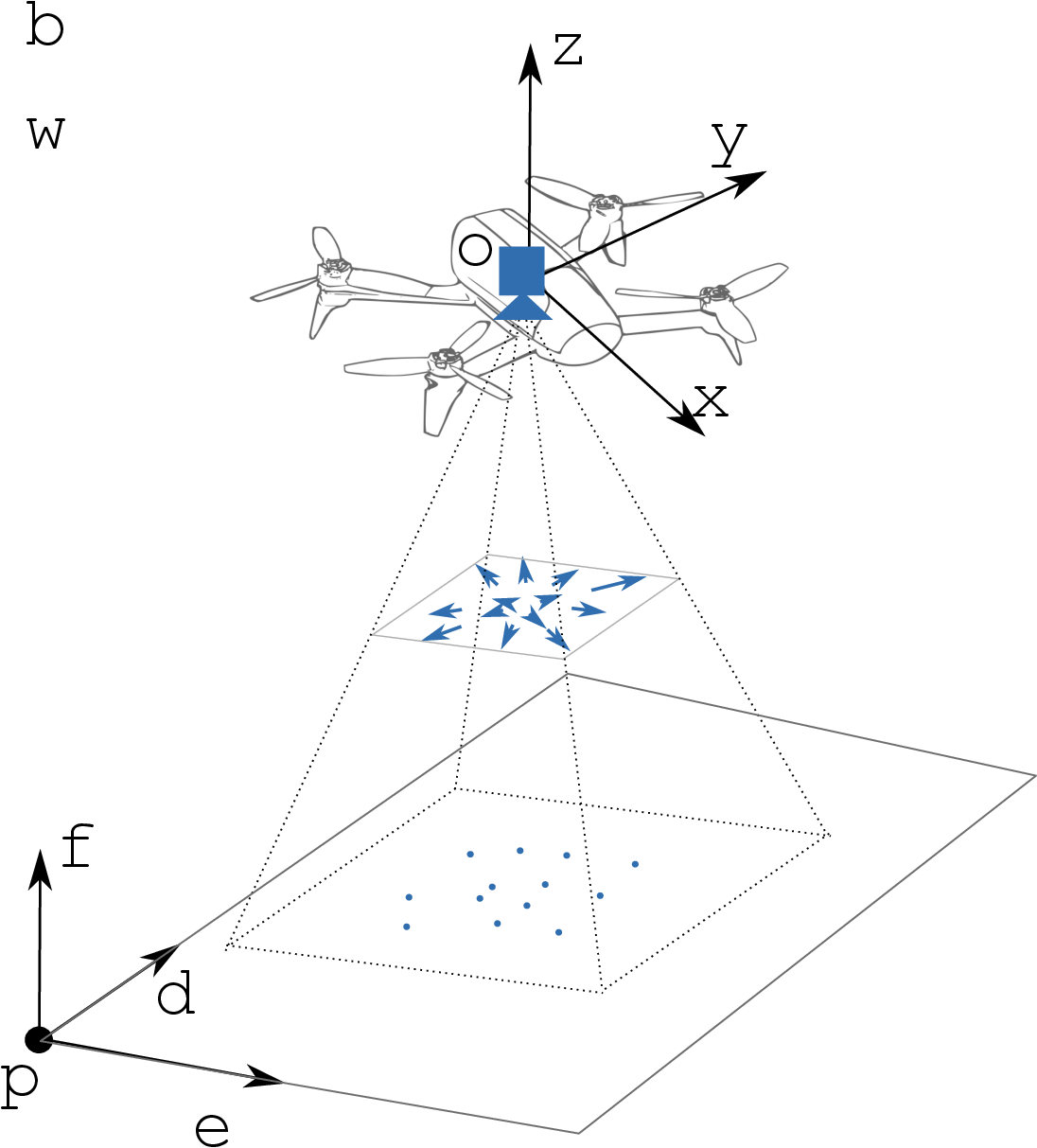}
\end{psfrags}%
%

%% file: INDI_G.tex
%
\providecommand\matlabfragNegTickNoWidth{\makebox[0pt][r]{\ensuremath{-}}}%
%
%
\providecommand\matlabtextA{\color[rgb]{0.150,0.150,0.150}\fontsize{7.00}{8.80}\selectfont\strut}%
\psfrag{063}[tc][tc]{\matlabtextA $t(s)$}%
\psfrag{065}[tc][tc]{\matlabtextA $t(s)$}%
\psfrag{067}[tc][tc]{\matlabtextA $t(s)$}%
\psfrag{068}[bc][tc]{\matlabtextA $diag(\hat{G})$}%
\psfrag{069}[bc][tc]{\matlabtextA $\bm{u}~(^o, ^o, N)$}%
\psfrag{070}[bc][tc]{\matlabtextA $\bm{\vartheta}~(rad/s)$}%
\psfrag{072}[bc][tc]{\matlabtextA $\bm{d}~(m)$}%
\providecommand\matlabtextB{\color[rgb]{0.000,0.000,0.000}\fontsize{7.50}{8.80}\selectfont\strut}%
\psfrag{064}[bc][tc]{\matlabtextB (c) $Z$-axis}%
\psfrag{066}[bc][tc]{\matlabtextB (b) $Y$-axis}%
\psfrag{071}[bc][tc]{\matlabtextB (a) $X$-axis}%
%
%
%
\providecommand\matlabtextC{\color[rgb]{0.150,0.150,0.150}\fontsize{7.00}{8.00}\selectfont\strut}%
\psfrag{000}[ct][ct]{\matlabtextC $0$}%
\psfrag{001}[ct][ct]{\matlabtextC $40$}%
\psfrag{005}[ct][ct]{\matlabtextC $0$}%
\psfrag{006}[ct][ct]{\matlabtextC $40$}%
\psfrag{011}[ct][ct]{\matlabtextC $0$}%
\psfrag{012}[ct][ct]{\matlabtextC $40$}%
\psfrag{017}[ct][ct]{\matlabtextC $0$}%
\psfrag{018}[ct][ct]{\matlabtextC $40$}%
\psfrag{022}[ct][ct]{\matlabtextC $0$}%
\psfrag{023}[ct][ct]{\matlabtextC $40$}%
\psfrag{028}[ct][ct]{\matlabtextC $0$}%
\psfrag{029}[ct][ct]{\matlabtextC $40$}%
\psfrag{033}[ct][ct]{\matlabtextC $0$}%
\psfrag{034}[ct][ct]{\matlabtextC $40$}%
\psfrag{039}[ct][ct]{\matlabtextC $0$}%
\psfrag{040}[ct][ct]{\matlabtextC $40$}%
\psfrag{045}[ct][ct]{\matlabtextC $0$}%
\psfrag{046}[ct][ct]{\matlabtextC $40$}%
\psfrag{050}[ct][ct]{\matlabtextC $0$}%
\psfrag{051}[ct][ct]{\matlabtextC $40$}%
\psfrag{055}[ct][ct]{\matlabtextC $0$}%
\psfrag{056}[ct][ct]{\matlabtextC $40$}%
\psfrag{059}[ct][ct]{\matlabtextC $0$}%
\psfrag{060}[ct][ct]{\matlabtextC $40$}%
%
%
%
\psfrag{002}[rc][rc]{\matlabtextC $-2$}%
\psfrag{003}[rc][rc]{\matlabtextC $0$}%
\psfrag{004}[rc][rc]{\matlabtextC $2$}%
\psfrag{007}[rc][rc]{\matlabtextC $-0.6$}%
\psfrag{008}[rc][rc]{\matlabtextC $-0.4$}%
\psfrag{009}[rc][rc]{\matlabtextC $-0.2$}%
\psfrag{010}[rc][rc]{\matlabtextC $0$}%
\psfrag{013}[rc][rc]{\matlabtextC $-0.1$}%
\psfrag{014}[rc][rc]{\matlabtextC $-0.05$}%
\psfrag{015}[rc][rc]{\matlabtextC $0$}%
\psfrag{016}[rc][rc]{\matlabtextC $0.05$}%
\psfrag{019}[rc][rc]{\matlabtextC $-200$}%
\psfrag{020}[rc][rc]{\matlabtextC $0$}%
\psfrag{021}[rc][rc]{\matlabtextC $200$}%
\psfrag{024}[rc][rc]{\matlabtextC $-0.5$}%
\psfrag{025}[rc][rc]{\matlabtextC $0$}%
\psfrag{026}[rc][rc]{\matlabtextC $0.5$}%
\psfrag{027}[rc][rc]{\matlabtextC $1$}%
\psfrag{030}[rc][rc]{\matlabtextC $0$}%
\psfrag{031}[rc][rc]{\matlabtextC $1$}%
\psfrag{032}[rc][rc]{\matlabtextC $2$}%
\psfrag{035}[rc][rc]{\matlabtextC $-6$}%
\psfrag{036}[rc][rc]{\matlabtextC $-4$}%
\psfrag{037}[rc][rc]{\matlabtextC $-2$}%
\psfrag{038}[rc][rc]{\matlabtextC $0$}%
\psfrag{041}[rc][rc]{\matlabtextC $-100$}%
\psfrag{042}[rc][rc]{\matlabtextC $-50$}%
\psfrag{043}[rc][rc]{\matlabtextC $0$}%
\psfrag{044}[rc][rc]{\matlabtextC $50$}%
\psfrag{047}[rc][rc]{\matlabtextC $-10$}%
\psfrag{048}[rc][rc]{\matlabtextC $-5$}%
\psfrag{049}[rc][rc]{\matlabtextC $0$}%
\psfrag{052}[rc][rc]{\matlabtextC $-200$}%
\psfrag{053}[rc][rc]{\matlabtextC $0$}%
\psfrag{054}[rc][rc]{\matlabtextC $200$}%
\psfrag{057}[rc][rc]{\matlabtextC $0$}%
\psfrag{058}[rc][rc]{\matlabtextC $2$}%
\psfrag{061}[rc][rc]{\matlabtextC $10$}%
\psfrag{062}[rc][rc]{\matlabtextC $15$}%
%
\fbox{\includegraphics[trim=20 0 30 -10, clip, width =0.48\textwidth ]{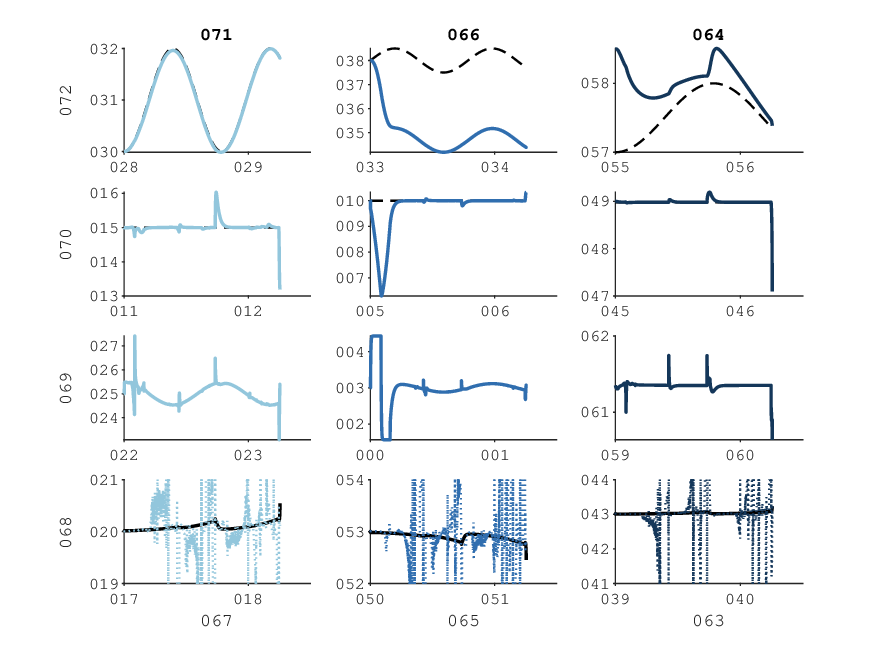}}

%% file: INDI_Ginv.tex
%
\providecommand\matlabfragNegTickNoWidth{\makebox[0pt][r]{\ensuremath{-}}}%
%
%
\providecommand\matlabtextA{\color[rgb]{0.150,0.150,0.150}\fontsize{7.00}{8.80}\selectfont\strut}%
\psfrag{064}[tc][tc]{\matlabtextA $t(s)$}%
\psfrag{066}[tc][tc]{\matlabtextA $t(s)$}%
\psfrag{068}[tc][tc]{\matlabtextA $t(s)$}%
\psfrag{069}[bc][bc]{\matlabtextA $diag(\hat{G}^\dagger)$}%
\psfrag{070}[bc][bc]{\matlabtextA $\bm{u}~(^o, ^o, N)$}%
\psfrag{071}[bc][bc]{\matlabtextA $\bm{\vartheta}~(rad/s)$}%
\psfrag{073}[bc][bc]{\matlabtextA $\bm{d}~(m)$}%
\providecommand\matlabtextB{\color[rgb]{0.000,0.000,0.000}\fontsize{7.50}{8.80}\selectfont\strut}%
\psfrag{065}[bc][tc]{\matlabtextB (c) $Z$-axis}%
\psfrag{067}[bc][tc]{\matlabtextB (b) $Y$-axis}%
\psfrag{072}[bc][tc]{\matlabtextB (a) $X$-axis}%
%
%
%
\providecommand\matlabtextC{\color[rgb]{0.150,0.150,0.150}\fontsize{7.00}{8.00}\selectfont\strut}%
\psfrag{000}[ct][ct]{\matlabtextC $0$}%
\psfrag{001}[ct][ct]{\matlabtextC $40$}%
\psfrag{005}[ct][ct]{\matlabtextC $0$}%
\psfrag{006}[ct][ct]{\matlabtextC $40$}%
\psfrag{011}[ct][ct]{\matlabtextC $0$}%
\psfrag{012}[ct][ct]{\matlabtextC $40$}%
\psfrag{017}[ct][ct]{\matlabtextC $0$}%
\psfrag{018}[ct][ct]{\matlabtextC $40$}%
\psfrag{022}[ct][ct]{\matlabtextC $0$}%
\psfrag{023}[ct][ct]{\matlabtextC $40$}%
\psfrag{028}[ct][ct]{\matlabtextC $0$}%
\psfrag{029}[ct][ct]{\matlabtextC $40$}%
\psfrag{033}[ct][ct]{\matlabtextC $0$}%
\psfrag{034}[ct][ct]{\matlabtextC $40$}%
\psfrag{038}[ct][ct]{\matlabtextC $0$}%
\psfrag{039}[ct][ct]{\matlabtextC $40$}%
\psfrag{044}[ct][ct]{\matlabtextC $0$}%
\psfrag{045}[ct][ct]{\matlabtextC $40$}%
\psfrag{049}[ct][ct]{\matlabtextC $0$}%
\psfrag{050}[ct][ct]{\matlabtextC $40$}%
\psfrag{054}[ct][ct]{\matlabtextC $0$}%
\psfrag{055}[ct][ct]{\matlabtextC $40$}%
\psfrag{059}[ct][ct]{\matlabtextC $0$}%
\psfrag{060}[ct][ct]{\matlabtextC $40$}%
%
%
%
\psfrag{002}[rc][rc]{\matlabtextC $-2$}%
\psfrag{003}[rc][rc]{\matlabtextC $-1$}%
\psfrag{004}[rc][rc]{\matlabtextC $0$}%
\psfrag{007}[rc][rc]{\matlabtextC $-4$}%
\psfrag{008}[rc][rc]{\matlabtextC $-2$}%
\psfrag{009}[rc][rc]{\matlabtextC $0$}%
\psfrag{010}[rc][rc]{\matlabtextC $2$}%
\psfrag{013}[rc][rc]{\matlabtextC $-0.1$}%
\psfrag{014}[rc][rc]{\matlabtextC }%
\psfrag{015}[rc][rc]{\matlabtextC $0$}%
\psfrag{016}[rc][rc]{\matlabtextC }%
\psfrag{019}[rc][rc]{\matlabtextC $-1$}%
\psfrag{020}[rc][rc]{\matlabtextC $-0.5$}%
\psfrag{021}[rc][rc]{\matlabtextC $0$}%
\psfrag{024}[rc][rc]{\matlabtextC $0$}%
\psfrag{025}[rc][rc]{\matlabtextC $1$}%
\psfrag{026}[rc][rc]{\matlabtextC $2$}%
\psfrag{027}[rc][rc]{\matlabtextC $3$}%
\psfrag{030}[rc][rc]{\matlabtextC $-0.2$}%
\psfrag{031}[rc][rc]{\matlabtextC $0$}%
\psfrag{032}[rc][rc]{\matlabtextC $0.2$}%
\psfrag{035}[rc][rc]{\matlabtextC $-10$}%
\psfrag{036}[rc][rc]{\matlabtextC $0$}%
\psfrag{037}[rc][rc]{\matlabtextC $10$}%
\psfrag{040}[rc][rc]{\matlabtextC $-4$}%
\psfrag{041}[rc][rc]{\matlabtextC $-2$}%
\psfrag{042}[rc][rc]{\matlabtextC $0$}%
\psfrag{043}[rc][rc]{\matlabtextC $2$}%
\psfrag{046}[rc][rc]{\matlabtextC $11.6$}%
\psfrag{047}[rc][rc]{\matlabtextC }%
\psfrag{048}[rc][rc]{\matlabtextC $11.8$}%
\psfrag{051}[rc][rc]{\matlabtextC $0$}%
\psfrag{052}[rc][rc]{\matlabtextC $1$}%
\psfrag{053}[rc][rc]{\matlabtextC $2$}%
\psfrag{056}[rc][rc]{\matlabtextC $0$}%
\psfrag{057}[rc][rc]{\matlabtextC }%
\psfrag{058}[rc][rc]{\matlabtextC $0.1$}%
\psfrag{061}[rc][rc]{\matlabtextC $-0.1$}%
\psfrag{062}[rc][rc]{\matlabtextC }%
\psfrag{063}[rc][rc]{\matlabtextC $0$}%
%
\fbox{\includegraphics[trim=20 0 30 -10, clip, width =0.48\textwidth ]{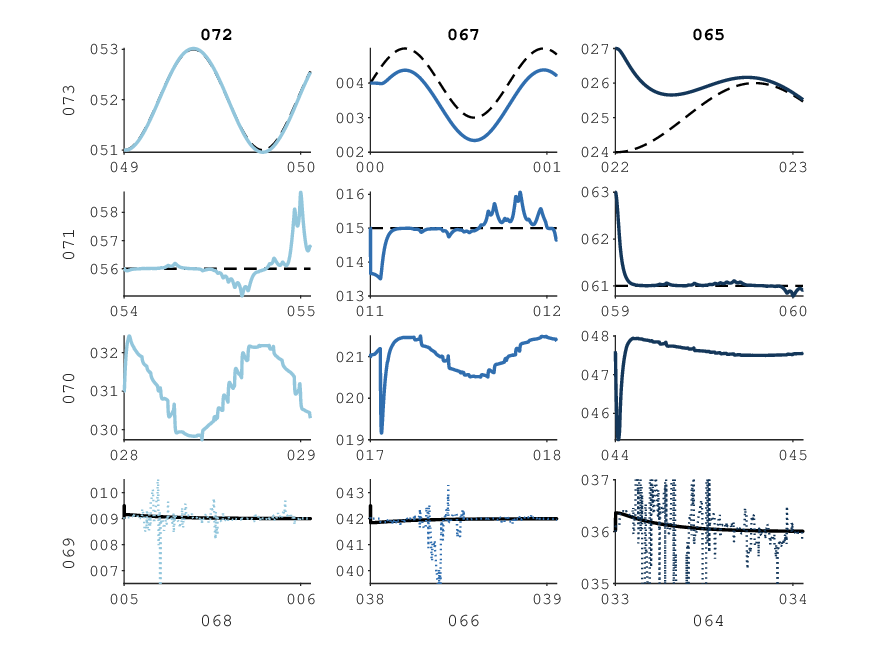}}

%% file: HardwareSoftware.tex
%
%
\begin{psfrags}%
\psfragscanon%
\newcommand{\tsize}{0.7}
%
\psfrag{a}[c][c][\tsize]{\color[rgb]{0,0,0}\setlength{\tabcolsep}{0pt}\begin{tabular}{c}On-board the MAV\end{tabular}}%
\psfrag{b}[c][c][\tsize]{\color[rgb]{0,0,0}\setlength{\tabcolsep}{0pt}\begin{tabular}{c}Vertical~Control\end{tabular}}%
\psfrag{c}[c][c][\tsize]{\color[rgb]{0,0,0}\setlength{\tabcolsep}{0pt}\begin{tabular}{c}Horizontal~Control\end{tabular}}%
\psfrag{d}[c][c][\tsize]{\color[rgb]{0,0,0}\setlength{\tabcolsep}{0pt}\begin{tabular}{c}Images\end{tabular}}%
\psfrag{e}[c][c][\tsize]{\color[rgb]{0,0,0}\setlength{\tabcolsep}{0pt}\begin{tabular}{c}FAST\\Feature\\ Detection\end{tabular}}%
\psfrag{f}[c][c][\tsize]{\color[rgb]{0,0,0}\setlength{\tabcolsep}{0pt}\begin{tabular}{c}$\bm{\mu_x}$,~$\bm{\mu_y}$\end{tabular}}%
\psfrag{g}[c][c][\tsize]{\color[rgb]{0,0,0}\setlength{\tabcolsep}{0pt}\begin{tabular}{c}Lucas-Kanade\\Feature\\Tracking\end{tabular}}%
\psfrag{h}[c][c][\tsize]{\color[rgb]{0,0,0}\setlength{\tabcolsep}{0pt}\begin{tabular}{c}$\bm{\mu_x}$,~$\bm{\mu_y}$,\\$\bm{\upsilon}_x$,~$\bm{\upsilon}_y$\end{tabular}}%
\psfrag{i}[c][c][\tsize]{\color[rgb]{0,0,0}\setlength{\tabcolsep}{0pt}\begin{tabular}{c}Optical Flow\\Estimation\end{tabular}}%
\psfrag{j}[c][c][\tsize]{\color[rgb]{0,0,0}\setlength{\tabcolsep}{0pt}\begin{tabular}{c}$\widehat{\vartheta}_x$, $\widehat{\vartheta}_y$,\\ $\widehat{\vartheta}_z$\end{tabular}}%
\psfrag{k}[c][c][\tsize]{\color[rgb]{0,0,0}\setlength{\tabcolsep}{0pt}\begin{tabular}{c}INDI\\Control\end{tabular}}%
\psfrag{l}[c][c][\tsize]{\color[rgb]{0,0,0}\setlength{\tabcolsep}{0pt}\begin{tabular}{c}$\bm{u}$\end{tabular}}%
\psfrag{m}[c][c][\tsize]{\color[rgb]{0,0,0}\setlength{\tabcolsep}{0pt}\begin{tabular}{c}Computer Vision Module\end{tabular}}%
\psfrag{n}[c][c][\tsize]{\color[rgb]{0,0,0}\setlength{\tabcolsep}{0pt}\begin{tabular}{c}Outer Loop Control\end{tabular}}%
\psfrag{o}[c][c][\tsize]{\color[rgb]{0,0,0}\setlength{\tabcolsep}{0pt}\begin{tabular}{c}Attitude and\\ Throttle\\ Control\end{tabular}}%
\psfrag{p}[c][c][\tsize]{\color[rgb]{0,0,0}\setlength{\tabcolsep}{0pt}\begin{tabular}{c}Inner Loop Control\end{tabular}}%
\psfrag{q}[c][c][\tsize]{\color[rgb]{0,0,0}\setlength{\tabcolsep}{0pt}\begin{tabular}{c}Motor\\ Mixing\end{tabular}}%
%
\includegraphics[width=0.5\textwidth]{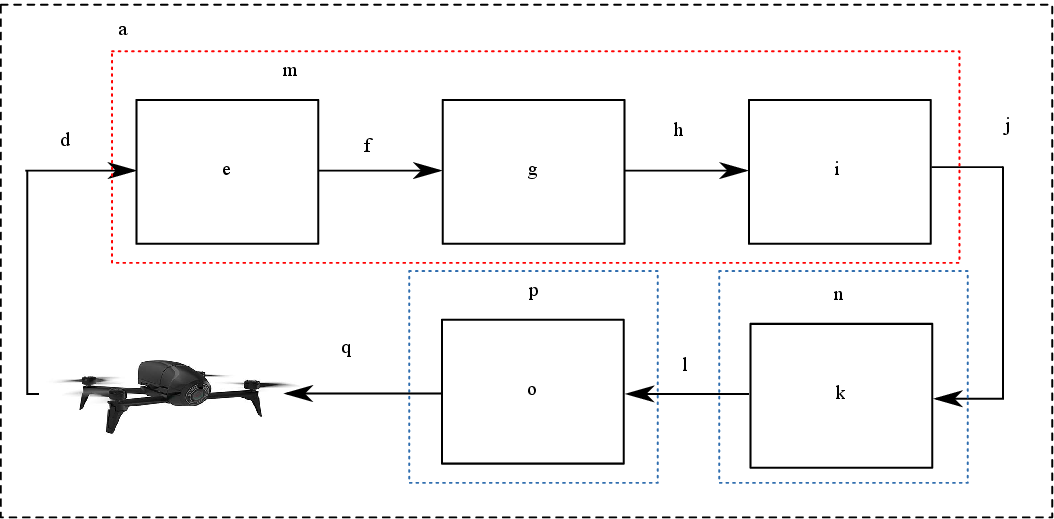}
\end{psfrags}%
%

%% file: landing_static_setpoints_inverseG.tex
%
\providecommand\matlabfragNegTickNoWidth{\makebox[0pt][r]{\ensuremath{-}}}%
%
%
\providecommand\matlabtextA{\color[rgb]{0.150,0.150,0.150}\fontsize{7.00}{8.80}\selectfont\strut}%
\psfrag{063}[tc][tc]{\matlabtextA $t(s)$}%
\psfrag{065}[tc][tc]{\matlabtextA $t(s)$}%
\psfrag{067}[tc][tc]{\matlabtextA $t(s)$}%
\psfrag{068}[bc][bc]{\matlabtextA $\widehat{G}^\dagger(\times~10^-3)$}%
\psfrag{069}[bc][bc]{\matlabtextA $\widehat{\vartheta}_z(rad/s)$}%
\psfrag{070}[bc][bc]{\matlabtextA $v_z(m/s)$}%
\psfrag{072}[bc][bc]{\matlabtextA $d_z(m)$}%
\providecommand\matlabtextB{\color[rgb]{0.000,0.000,0.000}\fontsize{8.00}{8.80}\bfseries\selectfont\strut}%
\psfrag{064}[bc][tc]{\matlabtextB (c) $\vartheta_z^*=-0.3~rad/s$}%
\psfrag{066}[bc][tc]{\matlabtextB (b) $\vartheta_z^*=-0.2~rad/s$}%
\psfrag{071}[bc][tc]{\matlabtextB (a) $\vartheta_z^*=-0.1~rad/s$}%
%
%
%
\providecommand\matlabtextC{\color[rgb]{0.150,0.150,0.150}\fontsize{8.00}{8.00}\selectfont\strut}%
\psfrag{000}[ct][ct]{\matlabtextC $0$}%
\psfrag{001}[ct][ct]{\matlabtextC $15$}%
\psfrag{005}[ct][ct]{\matlabtextC $0$}%
\psfrag{006}[ct][ct]{\matlabtextC $15$}%
\psfrag{010}[ct][ct]{\matlabtextC $0$}%
\psfrag{011}[ct][ct]{\matlabtextC $10$}%
\psfrag{015}[ct][ct]{\matlabtextC $0$}%
\psfrag{016}[ct][ct]{\matlabtextC $10$}%
\psfrag{020}[ct][ct]{\matlabtextC $0$}%
\psfrag{021}[ct][ct]{\matlabtextC $15$}%
\psfrag{026}[ct][ct]{\matlabtextC $0$}%
\psfrag{027}[ct][ct]{\matlabtextC $7$}%
\psfrag{032}[ct][ct]{\matlabtextC $0$}%
\psfrag{033}[ct][ct]{\matlabtextC $7$}%
\psfrag{037}[ct][ct]{\matlabtextC $0$}%
\psfrag{038}[ct][ct]{\matlabtextC $7$}%
\psfrag{041}[ct][ct]{\matlabtextC $0$}%
\psfrag{042}[ct][ct]{\matlabtextC $10$}%
\psfrag{047}[ct][ct]{\matlabtextC $0$}%
\psfrag{048}[ct][ct]{\matlabtextC $10$}%
\psfrag{053}[ct][ct]{\matlabtextC $0$}%
\psfrag{054}[ct][ct]{\matlabtextC $7$}%
\psfrag{058}[ct][ct]{\matlabtextC $0$}%
\psfrag{059}[ct][ct]{\matlabtextC $15$}%
%
%
%
\psfrag{017}[bc][bc]{}%
\psfrag{022}[bc][bc]{}%
\psfrag{055}[bc][bc]{}%
%
%
%
\psfrag{002}[rc][rc]{\matlabtextC $-0.4$}%
\psfrag{003}[rc][rc]{\matlabtextC $-0.2$}%
\psfrag{004}[rc][rc]{\matlabtextC $0$}%
\psfrag{007}[rc][rc]{\matlabtextC $-0.2$}%
\psfrag{008}[rc][rc]{\matlabtextC $-0.1$}%
\psfrag{009}[rc][rc]{\matlabtextC $0$}%
\psfrag{012}[rc][rc]{\matlabtextC $-0.4$}%
\psfrag{013}[rc][rc]{\matlabtextC $-0.2$}%
\psfrag{014}[rc][rc]{\matlabtextC $0$}%
\psfrag{018}[rc][rc]{\matlabtextC $0.5$}%
\psfrag{019}[rc][rc]{\matlabtextC $1$}%
\psfrag{023}[rc][rc]{\matlabtextC $0$}%
\psfrag{024}[rc][rc]{\matlabtextC $0.5$}%
\psfrag{025}[rc][rc]{\matlabtextC $1$}%
\psfrag{028}[rc][rc]{\matlabtextC $0$}%
\psfrag{029}[rc][rc]{\matlabtextC $1$}%
\psfrag{030}[rc][rc]{\matlabtextC $2$}%
\psfrag{031}[rc][rc]{\matlabtextC $3$}%
\psfrag{034}[rc][rc]{\matlabtextC $-1$}%
\psfrag{035}[rc][rc]{\matlabtextC $-0.5$}%
\psfrag{036}[rc][rc]{\matlabtextC $0$}%
\psfrag{039}[rc][rc]{\matlabtextC $-0.5$}%
\psfrag{040}[rc][rc]{\matlabtextC $0$}%
\psfrag{043}[rc][rc]{\matlabtextC $0$}%
\psfrag{044}[rc][rc]{\matlabtextC $1$}%
\psfrag{045}[rc][rc]{\matlabtextC $2$}%
\psfrag{046}[rc][rc]{\matlabtextC $3$}%
\psfrag{049}[rc][rc]{\matlabtextC $-0.6$}%
\psfrag{050}[rc][rc]{\matlabtextC $-0.4$}%
\psfrag{051}[rc][rc]{\matlabtextC $-0.2$}%
\psfrag{052}[rc][rc]{\matlabtextC $0$}%
\psfrag{056}[rc][rc]{\matlabtextC $0.5$}%
\psfrag{057}[rc][rc]{\matlabtextC $1$}%
\psfrag{060}[rc][rc]{\matlabtextC $0$}%
\psfrag{061}[rc][rc]{\matlabtextC $1$}%
\psfrag{062}[rc][rc]{\matlabtextC $2$}%
%
\fbox{\includegraphics[trim=20 0 30 -10, clip, width =0.48\textwidth ]{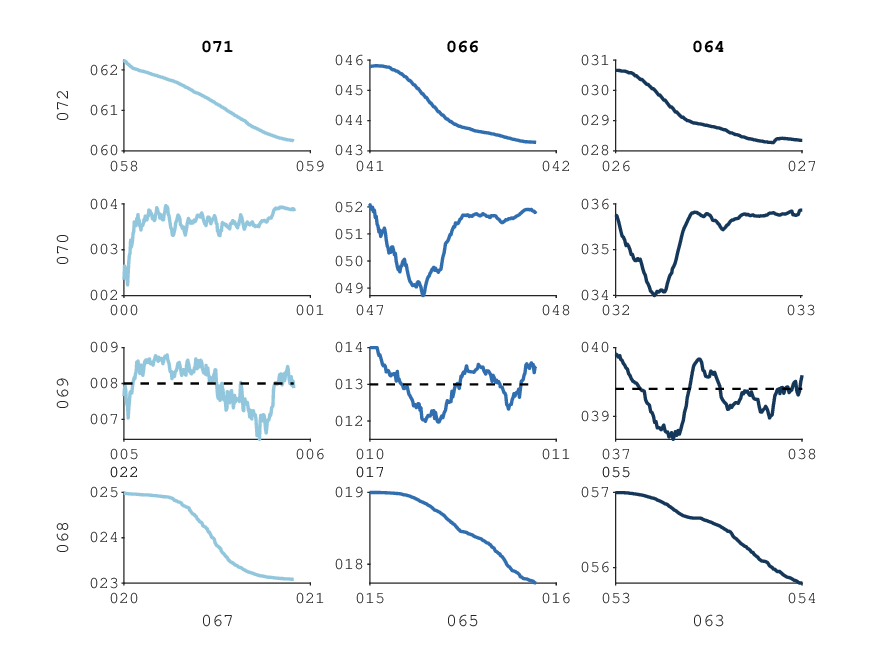}}

%% file: moving_platform4.tex
%
\providecommand\matlabfragNegTickNoWidth{\makebox[0pt][r]{\ensuremath{-}}}%
%
%
\providecommand\matlabtextA{\color[rgb]{0.000,0.000,0.000}\fontsize{7.65}{7.65}\selectfont\strut}%
\psfrag{00000}[cl][cl]{\matlabtextA relative}%
\psfrag{00001}[cl][cl]{\matlabtextA absolute}%
\providecommand\matlabtextB{\color[rgb]{0.150,0.150,0.150}\fontsize{9.35}{9.35}\selectfont\strut}%
\psfrag{061}[tc][tc]{\matlabtextB $t~(s)$}%
\psfrag{063}[tc][tc]{\matlabtextB $t~(s)$}%
\psfrag{065}[tc][tc]{\matlabtextB $t~(s)$}%
\psfrag{066}[bc][bc]{\matlabtextB $\widehat{\vartheta}_y~(rad/s)$}%
\psfrag{067}[bc][bc]{\matlabtextB $\widehat{\vartheta}_z~(rad/s)$}%
\psfrag{069}[bc][bc]{\matlabtextB $d_z~(m)$}%
\providecommand\matlabtextC{\color[rgb]{0.000,0.000,0.000}\fontsize{8.50}{9.35}\selectfont\strut}%
\psfrag{062}[bc][tc]{\matlabtextC (c) $\vartheta_z^*=-0.1~rad/s$, $\vartheta_y^*=-0.8~rad/s$}%
\psfrag{064}[bc][tc]{\matlabtextC (b) $\vartheta_z^*=-0.2~rad/s$, $\vartheta_y^*=0~rad/s$}%
\psfrag{068}[bc][tc]{\matlabtextC (a) $\vartheta_z^*=0~rad/s$, $\vartheta_y^*=0~rad/s$}%
%
%
%
\providecommand\matlabtextD{\color[rgb]{0.150,0.150,0.150}\fontsize{8.50}{8.50}\selectfont\strut}%
\psfrag{002}[ct][ct]{\matlabtextD $0$}%
\psfrag{003}[ct][ct]{\matlabtextD $5$}%
\psfrag{004}[ct][ct]{\matlabtextD $10$}%
\psfrag{008}[ct][ct]{\matlabtextD $0$}%
\psfrag{009}[ct][ct]{\matlabtextD $5$}%
\psfrag{010}[ct][ct]{\matlabtextD $10$}%
\psfrag{015}[ct][ct]{\matlabtextD $0$}%
\psfrag{016}[ct][ct]{\matlabtextD $20$}%
\psfrag{017}[ct][ct]{\matlabtextD $40$}%
\psfrag{022}[ct][ct]{\matlabtextD $0$}%
\psfrag{023}[ct][ct]{\matlabtextD $5$}%
\psfrag{024}[ct][ct]{\matlabtextD $10$}%
\psfrag{028}[ct][ct]{\matlabtextD $0$}%
\psfrag{029}[ct][ct]{\matlabtextD $5$}%
\psfrag{030}[ct][ct]{\matlabtextD $10$}%
\psfrag{035}[ct][ct]{\matlabtextD $0$}%
\psfrag{036}[ct][ct]{\matlabtextD $20$}%
\psfrag{037}[ct][ct]{\matlabtextD $40$}%
\psfrag{041}[ct][ct]{\matlabtextD $0$}%
\psfrag{042}[ct][ct]{\matlabtextD $20$}%
\psfrag{043}[ct][ct]{\matlabtextD $40$}%
\psfrag{047}[ct][ct]{\matlabtextD $0$}%
\psfrag{048}[ct][ct]{\matlabtextD $5$}%
\psfrag{049}[ct][ct]{\matlabtextD $10$}%
\psfrag{055}[ct][ct]{\matlabtextD $0$}%
\psfrag{056}[ct][ct]{\matlabtextD $5$}%
\psfrag{057}[ct][ct]{\matlabtextD $10$}%
%
%
%
\psfrag{005}[rc][rc]{\matlabtextD $-0.5$}%
\psfrag{006}[rc][rc]{\matlabtextD $0$}%
\psfrag{007}[rc][rc]{\matlabtextD $0.5$}%
\psfrag{011}[rc][rc]{\matlabtextD $0$}%
\psfrag{012}[rc][rc]{\matlabtextD $0.5$}%
\psfrag{013}[rc][rc]{\matlabtextD $1$}%
\psfrag{014}[rc][rc]{\matlabtextD $1.5$}%
\psfrag{018}[rc][rc]{\matlabtextD $0$}%
\psfrag{019}[rc][rc]{\matlabtextD $0.5$}%
\psfrag{020}[rc][rc]{\matlabtextD $1$}%
\psfrag{021}[rc][rc]{\matlabtextD $1.5$}%
\psfrag{025}[rc][rc]{\matlabtextD $-2$}%
\psfrag{026}[rc][rc]{\matlabtextD $-1$}%
\psfrag{027}[rc][rc]{\matlabtextD $0$}%
\psfrag{031}[rc][rc]{\matlabtextD $-0.4$}%
\psfrag{032}[rc][rc]{\matlabtextD $-0.2$}%
\psfrag{033}[rc][rc]{\matlabtextD $0$}%
\psfrag{034}[rc][rc]{\matlabtextD $0.2$}%
\psfrag{038}[rc][rc]{\matlabtextD $-0.5$}%
\psfrag{039}[rc][rc]{\matlabtextD $0$}%
\psfrag{040}[rc][rc]{\matlabtextD $0.5$}%
\psfrag{044}[rc][rc]{\matlabtextD $-0.2$}%
\psfrag{045}[rc][rc]{\matlabtextD $0$}%
\psfrag{046}[rc][rc]{\matlabtextD $0.2$}%
\psfrag{050}[rc][rc]{\matlabtextD $-0.6$}%
\psfrag{051}[rc][rc]{\matlabtextD $-0.4$}%
\psfrag{052}[rc][rc]{\matlabtextD $-0.2$}%
\psfrag{053}[rc][rc]{\matlabtextD $0$}%
\psfrag{054}[rc][rc]{\matlabtextD $0.2$}%
\psfrag{058}[rc][rc]{\matlabtextD $0$}%
\psfrag{059}[rc][rc]{\matlabtextD $1$}%
\psfrag{060}[rc][rc]{\matlabtextD $2$}%
%
 \begin{framed}
\centering
\begin{subfigure}[b]{1.0\textwidth}
     \centering
     \includegraphics[trim=0 0 0 0, clip, width =0.9\textwidth]{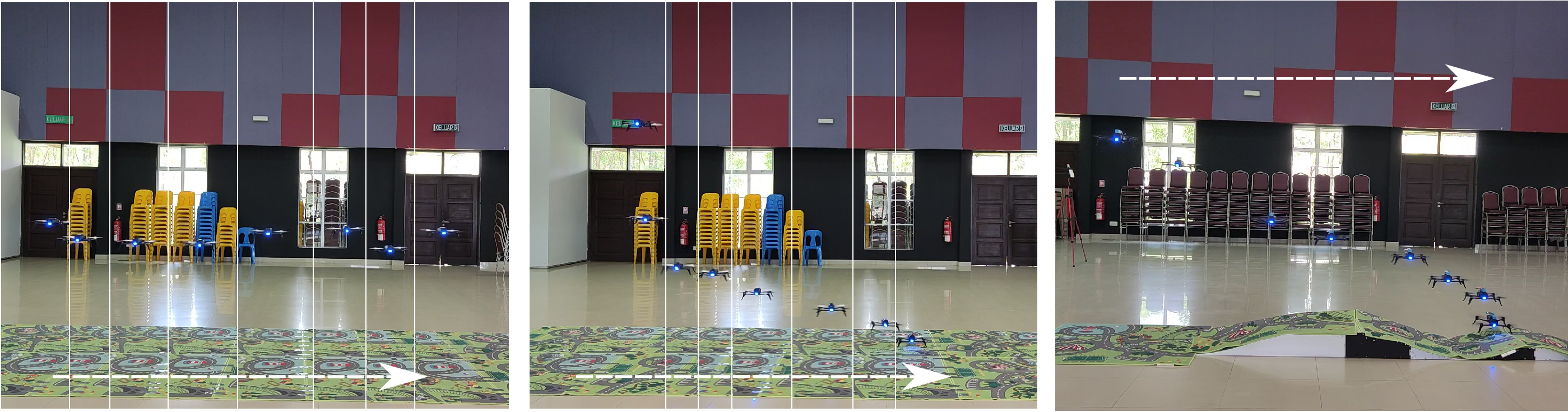}
     \label{fig:y equals x}
 \end{subfigure} 
 \hfill
 \begin{subfigure}[b]{1.0\textwidth}
     \centering
     \includegraphics[trim=55 0 40 0, clip, width =0.85\textwidth, height =0.65\textwidth ]{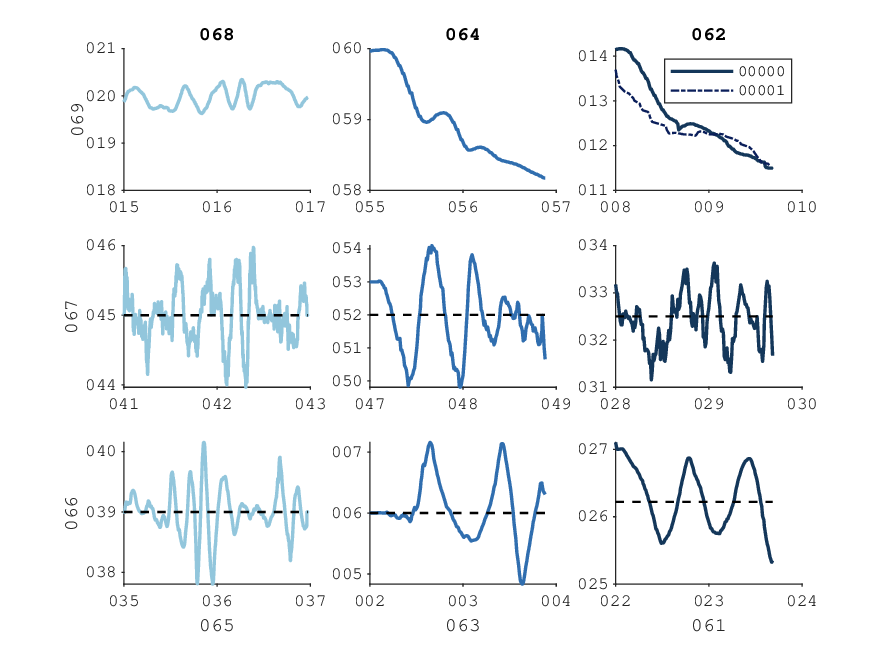} 
     \label{fig:three sin x}
 \end{subfigure}
\end{framed}

%% file: paper_template.bbl
\begin{thebibliography}{21}
\providecommand{\natexlab}[1]{#1}
\providecommand{\url}[1]{\texttt{#1}}
\expandafter\ifx\csname urlstyle\endcsname\relax
  \providecommand{\doi}[1]{doi: #1}\else
  \providecommand{\doi}{doi: \begingroup \urlstyle{rm}\Url}\fi

\bibitem[Soria(2022)]{soria2022swarms}
Enrica Soria.
\newblock Swarms of flying robots in unknown environments.
\newblock \emph{Science Robotics}, 7\penalty0 (66):\penalty0 eabq2215, 2022.

\bibitem[Fridovich-Keil et~al.(2020)Fridovich-Keil, Bajcsy, Fisac, Herbert,
  Wang, Dragan, and Tomlin]{fridovich2020confidence}
David Fridovich-Keil, Andrea Bajcsy, Jaime~F Fisac, Sylvia~L Herbert, Steven
  Wang, Anca~D Dragan, and Claire~J Tomlin.
\newblock Confidence-aware motion prediction for real-time collision
  avoidance1.
\newblock \emph{The International Journal of Robotics Research}, 39\penalty0
  (2-3):\penalty0 250--265, 2020.

\bibitem[Wang et~al.(2022)Wang, Melfi~Jr, and Leonardo]{wang2022recovery}
Z~Jane Wang, James Melfi~Jr, and Anthony Leonardo.
\newblock Recovery mechanisms in the dragonfly righting reflex.
\newblock \emph{Science}, 376\penalty0 (6594):\penalty0 754--758, 2022.

\bibitem[Collett(2002)]{collett2002insect}
Thomas~S Collett.
\newblock Insect vision: controlling actions through optic flow.
\newblock \emph{Current Biology}, 12\penalty0 (18):\penalty0 R615--R617, 2002.

\bibitem[Yu et~al.(2022)Yu, Zardini, Censi, and Fuller]{yu2022visual}
Zhitao Yu, Gioele Zardini, Andrea Censi, and Sawyer Fuller.
\newblock Visual confined-space navigation using an efficient learned bilinear
  optic flow approximation for insect-scale robots.
\newblock In \emph{2022 IEEE/RSJ International Conference on Intelligent Robots
  and Systems (IROS)}, pages 4250--4256. IEEE, 2022.

\bibitem[Mahlknecht et~al.(2022)Mahlknecht, Gehrig, Nash, Rockenbauer, Morrell,
  Delaune, and Scaramuzza]{mahlknecht2022exploring}
Florian Mahlknecht, Daniel Gehrig, Jeremy Nash, Friedrich~M Rockenbauer,
  Benjamin Morrell, Jeff Delaune, and Davide Scaramuzza.
\newblock Exploring event camera-based odometry for planetary robots.
\newblock \emph{IEEE Robotics and Automation Letters}, 7\penalty0 (4):\penalty0
  8651--8658, 2022.

\bibitem[Falanga et~al.(2020)Falanga, Kleber, and
  Scaramuzza]{falanga2020dynamic}
Davide Falanga, Kevin Kleber, and Davide Scaramuzza.
\newblock Dynamic obstacle avoidance for quadrotors with event cameras.
\newblock \emph{Science Robotics}, 5\penalty0 (40):\penalty0 eaaz9712, 2020.

\bibitem[Ruffier and Franceschini(2015)]{ruffier2015optic}
Franck Ruffier and Nicolas Franceschini.
\newblock Optic flow regulation in unsteady environments: A tethered mav
  achieves terrain following and targeted landing over a moving platform.
\newblock \emph{Journal of Intelligent \& Robotic Systems}, 79:\penalty0
  275--293, 2015.

\bibitem[Heriss{\'e} et~al.(2011)Heriss{\'e}, Hamel, Mahony, and
  Russotto]{herisse2011landing}
Bruno Heriss{\'e}, Tarek Hamel, Robert Mahony, and Fran{\c{c}}ois-Xavier
  Russotto.
\newblock Landing a vtol unmanned aerial vehicle on a moving platform using
  optical flow.
\newblock \emph{IEEE Transactions on robotics}, 28\penalty0 (1):\penalty0
  77--89, 2011.

\bibitem[Ho et~al.(2018)Ho, de~Croon, van Kampen, Chu, and
  Mulder]{ho2018adaptive}
H~W Ho, G~C H~E de~Croon, E~van Kampen, QP~Chu, and Max Mulder.
\newblock Adaptive gain control strategy for constant optical flow divergence
  landing.
\newblock \emph{IEEE Transactions on Robotics}, 34\penalty0 (2):\penalty0
  508--516, 2018.

\bibitem[De~Croon et~al.(2022)De~Croon, Dupeyroux, De~Wagter, Chatterjee,
  Olejnik, and Ruffier]{de2022accommodating}
Guido~CHE De~Croon, Julien~JG Dupeyroux, Christophe De~Wagter, Abhishek
  Chatterjee, Diana~A Olejnik, and Franck Ruffier.
\newblock Accommodating unobservability to control flight attitude with optic
  flow.
\newblock \emph{Nature}, 610\penalty0 (7932):\penalty0 485--490, 2022.

\bibitem[O’Connell et~al.(2022)O’Connell, Shi, Shi, Azizzadenesheli,
  Anandkumar, Yue, and Chung]{o2022neural}
Michael O’Connell, Guanya Shi, Xichen Shi, Kamyar Azizzadenesheli, Anima
  Anandkumar, Yisong Yue, and Soon-Jo Chung.
\newblock Neural-fly enables rapid learning for agile flight in strong winds.
\newblock \emph{Science Robotics}, 7\penalty0 (66):\penalty0 eabm6597, 2022.

\bibitem[Kendoul(2014)]{kendoul2014four}
Farid Kendoul.
\newblock Four-dimensional guidance and control of movement using
  time-to-contact: Application to automated docking and landing of unmanned
  rotorcraft systems.
\newblock \emph{The International Journal of Robotics Research}, 33\penalty0
  (2):\penalty0 237--267, 2014.

\bibitem[Cesetti et~al.(2010)Cesetti, Frontoni, Mancini, Zingaretti, and
  Longhi]{cesetti2010vision}
Andrea Cesetti, Emanuele Frontoni, Adriano Mancini, Primo Zingaretti, and Sauro
  Longhi.
\newblock A vision-based guidance system for uav navigation and safe landing
  using natural landmarks.
\newblock \emph{Journal of intelligent and robotic systems}, 57:\penalty0
  233--257, 2010.

\bibitem[Zhou et~al.(2021)Zhou, Ho, and Chu]{zhou2021extended}
Ye~Zhou, Hann~Woei Ho, and Qiping Chu.
\newblock Extended incremental nonlinear dynamic inversion for optical flow
  control of micro air vehicles.
\newblock \emph{Aerospace Science and Technology}, 116:\penalty0 106889, 2021.

\bibitem[Steffensen et~al.(2022)Steffensen, Steinert, and
  Smeur]{steffensen2022nonlinear}
Rasmus Steffensen, Agnes Steinert, and Ewoud~JJ Smeur.
\newblock Nonlinear dynamic inversion with actuator dynamics: An incremental
  control perspective.
\newblock \emph{Journal of Guidance, Control, and Dynamics}, pages 1--9, 2022.

\bibitem[Smeur et~al.(2018)Smeur, de~Croon, and Chu]{smeur2018cascaded}
Ewoud~JJ Smeur, Guido~CHE de~Croon, and Qiping Chu.
\newblock Cascaded incremental nonlinear dynamic inversion for mav disturbance
  rejection.
\newblock \emph{Control Engineering Practice}, 73:\penalty0 79--90, 2018.

\bibitem[van't Veld et~al.(2018)van't Veld, Van~Kampen, and
  Chu]{van2018stability}
Ronald van't Veld, Erik-Jan Van~Kampen, and Qiping Chu.
\newblock Stability and robustness analysis and improvements for incremental
  nonlinear dynamic inversion control.
\newblock In \emph{2018 AIAA Guidance, Navigation, and Control Conference},
  page 1127, 2018.

\bibitem[Rosten and Drummond(2006)]{rosten2006machine}
Edward Rosten and Tom Drummond.
\newblock Machine learning for high-speed corner detection.
\newblock In \emph{Computer Vision--ECCV 2006}, pages 430--443. Springer, 2006.

\bibitem[Bouquet(2000)]{bouquet2000pyramidal}
JY~Bouquet.
\newblock Pyramidal implementation of the {L}ucas {K}anade feature tracker.
\newblock \emph{Intel Corporation, Microprocessor Research Labs}, 2000.

\bibitem[Ching et~al.(2022)Ching, Tan, and Ho]{ching2022ultra}
Poh~Ling Ching, Shu~Chuan Tan, and Hann~Woei Ho.
\newblock Ultra-wideband localization and deep-learning-based plant monitoring
  using micro air vehicles.
\newblock \emph{Journal of Aerospace Information Systems}, 19\penalty0
  (11):\penalty0 717--728, 2022.

\end{thebibliography}
